\documentclass[preprint,12pt,authoryear]{elsarticle}

\usepackage{amssymb}
\usepackage{gensymb}
\usepackage{graphicx}
\usepackage{amsmath,amsfonts}
\usepackage{algorithm}
\usepackage{algpseudocode}
\usepackage{subcaption}
\usepackage{comment}
\usepackage{lscape}
\usepackage{hyperref}
\usepackage{multirow}
\usepackage{rotating}
\usepackage{forest}
\usepackage{listings}
\usepackage{nomencl}
\makenomenclature

\usepackage{xcolor}
\definecolor{myblue}{RGB}{0,0,0}
\usepackage{hyperref}
\DeclareMathOperator{\arcsinh}{sinh^{-1}}
\DeclareMathOperator{\arctanh}{tanh^{-1}}
\usepackage{multirow}
\hypersetup{
    colorlinks=true,
    linkcolor=blue,
    citecolor=blue,
    urlcolor=blue
}
\usepackage{pifont}
\newcommand{\cmark}{\ding{51}}%
\newcommand{\xmark}{\ding{55}}%
\usepackage{enumitem}
\lstdefinestyle{hierarchical}{
    basicstyle=\footnotesize\ttfamily,
    numbers=none,
    frame=none,
    breaklines=true,
    postbreak=\mbox{\textcolor{gray}{$\hookrightarrow$}\space},
    showstringspaces=false,
    keywordstyle=\bfseries\color{blue},
    keywords={If},
    emphstyle=\bfseries\color{purple},
    emph={[2]Choice:},
    commentstyle=\color{gray},
    emph={[3]Boolean},
    emphstyle={[3]\color{olive}},
    emph={[4]Choice:},
    emphstyle={[4]\color{magenta}},
    emph={[5]Parent condition:},
    emphstyle={[5]\color{teal}},
    tabsize=2,
    literate=*{--}{--}1
}

\lstdefinestyle{mystyle}{
    language=Python,
    style=hierarchical
}

\journal{Knowledge-Based Systems}

\begin{document}

\begin{frontmatter}



\title{Activation Function Optimization Scheme for Image Classification}


\author[inst1]{Abdur Rahman}

\affiliation[inst1]{organization={Department of Industrial and Systems Engineering},
            addressline={Mississippi State University}, 
            city={Mississippi State},
            postcode={MS 39762}, 
            country={USA}}
\author[inst2]{Lu He}

\author[inst1]{Haifeng Wang\corref{cor1}}
\cortext[cor1]{Corresponding author: Haifeng Wang (E-mail: wang@ise.msstate.edu)}

\affiliation[inst2]{organization={Department of Marketing, Quantitative Analysis, and Business Law},
            addressline={Mississippi State University}, 
            city={Mississippi State},
            postcode={MS 39762}, 
            country={USA}}

\begin{abstract}

Activation function has a significant impact on the dynamics, convergence, and performance of deep neural networks. The search for a consistent and high-performing activation function has always been a pursuit during deep learning model development. {\color{myblue}Existing state-of-the-art activation functions are manually designed with human expertise except for Swish. Swish was developed using a reinforcement learning-based search strategy. In this study, we propose an evolutionary approach for optimizing activation functions specifically for image classification tasks, aiming to discover functions that outperform current state-of-the-art options.} Through this optimization framework, we obtain a series of high{\color{myblue}-}performing activation functions denoted as Exponential Error Linear Unit (EELU). The {\color{myblue}developed} activation functions are {\color{myblue}evaluated} for image classification tasks from two perspectives: (1) five state-of-{\color{myblue}the-}art neural network architectures, such as ResNet50, AlexNet, VGG16, MobileNet, and Compact Convolutional Transformer which cover computationally heavy to light neural networks, and (2) eight standard datasets, including CIFAR10, Imagenette, MNIST, Fashion MNIST, Beans, Colorectal Histology, CottonWeedID15, and TinyImageNet which cover from typical machine vision benchmark, agricultural image applications to medical image applications. Finally, we statistically investigate the {\color{myblue}generalization} of the resultant activation functions developed through the optimization scheme. With a Friedman test, we conclude that the optimization scheme is able to generate activation functions that outperform the existing standard ones in {\color{myblue}92.8\% cases among 28 different cases studied}, and $-x\cdot erf(e^{-x})$ is found {\color{myblue}to be} the best activation function for image classification generated by the optimization scheme.
\end{abstract}








\begin{keyword}
Activation function \sep evolutionary approach \sep Exponential Error Linear Unit (EELU)\sep Genetic Algorithm
\end{keyword}

\end{frontmatter}






\section{Introduction}
\label{sec:introduction}
The choice of activation function plays a pivotal role in determining the learning dynamics, ability to converge, convergence speed, and ultimate performance of a deep neural network. Activation functions determine whether a neuron should be active or not during the training of a neural network \citep{zahedinasab2020neuroevolutionary}. Activation function can be divided into three types: (1) linear activation function, (2) nonlinear monotonic activation function, and (3) nonlinear non-monotonic activation function~\citep{zhu2021logish}. The first three linear activation functions are binary step function~\citep{sharma2017activation}, sign function~\citep{655045}, and identity function. Due to the poor classification ability of these functions, researchers have introduced some nonlinear monotonic functions such as Sigmoid~\citep{narayan1997generalized}, Tanh~\citep{kalman1992tanh}, ReLU~\citep{nair2010rectified,glorot2011deep}, and their variant functions~\citep{courbariaux2015binaryconnect,gulcehre2016noisy,arjovsky2016unitary}. Sigmoid and Tanh are the first two nonlinear functions used in deep neural networks~\citep{apicella2021survey}. However, both of them have the same limitation of gradient vanishing problem~\citep{hochreiter1998vanishing}. The emergence of ReLU has paved the way to a better activation function by addressing the gradient vanishing problem \citep{hu2018overcoming}. Though ReLU has a monotonically increasing part in the positive $x$-axis, it becomes saturated on the negative side, which suppresses every negative signal to zero and leads to significant information loss from the neural network. The saturation on the negative side is typically considered as the dying ReLU problem~\citep{lu2019dying} and {\color{myblue}Leaky ReLU (LReLU)}~\citep{maas2013rectifier} was thus proposed. {\color{myblue}LReLU} allows a small degree of negative input information by {\color{myblue}adding} a {\color{myblue}slight} {\color{myblue}slope} of 0.1. In addition to that, different variants of ReLU were proposed over the years{\color{myblue},} such as PReLU~\citep{he2015delving}, RReLU~\citep{xu2015empirical}, and SReLU~\citep{jin2016deep}. {\color{myblue}ELU (Exponential Linear Unit) ~\citep{clevert2015fast} is another extension of the ReLU, designed to address the ``dying ReLU" problem. ELU introduces an exponential term for negative inputs, which helps maintain a small, non-zero gradient and improves the learning process.} An update of ELU with self-normalizing properties was proposed as SELU~\citep{klambauer2017self}.

{\color{myblue}The} nonlinear non-monotonic functions are gaining more popularity for their well{\color{myblue}-}generalized performance {\color{myblue}across} different neural architectures and {\color{myblue}over} complex datasets. Swish~\citep{ramachandran2017searching}, GeLU~\citep{hendrycks2016gaussian}, and Mish~\citep{misra2019mish} are the most widely used non-monotonic activation functions. The log-Softplus ERror activation Function (SERF)~\citep{nag2021serf} and Logish~\citep{zhu2021logish} {\color{myblue}have been} proposed very recently{\color{myblue},} and they also perform well due to their non-monotonicity. {\color{myblue}These} non-monotonic functions have {\color{myblue}similar properties, making} them better performing than other linear and nonlinear monotonic functions. These functions are unbounded above ($x$-axis), bounded below, possess a non-monotonic region, demonstrate differentiability almost everywhere, and show a smooth curve at almost all points allowing better information flow through the neural network.

The search for an activation function that possesses the mentioned desirable properties and also outperforms the {\color{myblue}state-of-the-art} activation functions has been a point of interest for the researcher over the years. {\color{myblue}A comparative study of activation functions for convolutional neural networks (CNN) has been carried out by \cite{vargas2021activation}. The authors \citep{vargas2021activation} proposed two manually developed activation functions for CNN and compared their performance with existing activation functions using CIFAR10 \citep{Krizhevsky09learningmultiple}, CIFAR100 \citep{krizhevsky2009learning}, CINIC10 \citep{darlow2018cinic}, and ImageNet \citep{ILSVRC15} datasets. In another study, \cite{hu2022adaptively} proposed a parameterized activation function called AReLU to fit the parameters during model training adaptively. \cite{hu2022adaptively} compared the activation function across a range of datasets including CIFAR10 \citep{Krizhevsky09learningmultiple}, CIFAR100 \citep{krizhevsky2009learning}, miniImageNet \citep{shaban2017one}, PASCAL VOC \citep{Everingham15}, and COCO \citep{lin2014microsoft}. Task-specific activation function, e.g., a discontinuous hard-limiting activation function, was proposed by \cite{liu2008one} to solve quadratic programming problems. Except for the Swish, most activation functions described above are hand-crafted. The development of Swish using reinforcement learning has opened the door to search for better activation functions. The superiority of the performance of Swish over ReLU convinced us to take a step forward and search for a more versatile activation function}. {\color{myblue}Evolutionary}  algorithms have {\color{myblue}greatly succeeded} in optimizing deep neural network architectures {\color{myblue}in recent years}. This application is not limited to optimizing the architectures but also has significant potential in finding the most effective loss functions, the highest performing activation functions{\color{myblue},} and the optimal combination of hyper-parameters. Motivated by this, we proposed an evolutionary optimization{\color{myblue}-}based scheme to develop the highest{\color{myblue}-}performing activation function for image classification tasks. We also performed experiments to statistically compare the performance of the developed activation functions with the existing standard ones.

{\color{myblue}\cite{dubey2022activation} recently presented a comprehensive survey of activation functions and evaluated their performance on 6 different deep neural networks with CIFAR10 and CIFAR100 datasets. The key difference between our work and this study \citep{dubey2022activation} is that apart from studying the performance of state-of-the-art activation functions across different datasets and neural networks, we have proposed a series of high-performing activation functions for image classification tasks. While  \cite{dubey2022activation} only focused on CIFAR10 and CIFAR100, we compared the performance of these proposed activation functions with the existing state-of-the-art ones across eight different image classification datasets including two large-scale benchmark datasets, TinyImageNet (200 classes) \citep{le2015tiny}, and CottonWeedID15 (15 classes) \citep{chen2022performance}.}

The structure of this paper is organized as {\color{myblue}follows}. Section II provides an overview of related algorithms proposed to optimize activation functions. Section III describes the proposed activation function optimization scheme. Experimental results are demonstrated in Section IV. The research findings and future work are concluded in Section V.
\begin{table*}[h]
\centering
\caption{Summary of related works on the development of activation functions for image classification}
{\color{myblue}
\label{tab:table-a}
\resizebox{\linewidth}{!}{
\begin{tabular}{@{}ccccc@{}}
\hline
\textbf{Studies} &
  \textbf{\cite{basirat2018quest}} &
  \textbf{\cite{bingham2020evolutionary}} &
  \textbf{\cite{nader2021evolution}} &
  \textbf{Our Work} \\ \hline
\textbf{\begin{tabular}[c]{@{}c@{}}Functional Elements \\ in Search Space\end{tabular}} &
  \begin{tabular}[c]{@{}c@{}}Existing activation \\ functions\end{tabular} &
  \begin{tabular}[c]{@{}c@{}}Search space defined by\\  \cite{ramachandran2017searching}\end{tabular} &
  \begin{tabular}[c]{@{}c@{}}Existing activation function \\ and trigonometric functions\end{tabular} &
  \begin{tabular}[c]{@{}c@{}}Carefully chosen list of functions \\ from search space defined by \\  \cite{ramachandran2017searching}, \\ and several modified binary\\ functions of $b(u_1(x),u_2(x))$ form\end{tabular} \\\hline
\textbf{Fitness Function} &
  Test Accuracy based &
  \begin{tabular}[c]{@{}c@{}}Validation accuracy and \\ validation loss based\end{tabular} &
  Validation accuracy based &
  \begin{tabular}[c]{@{}c@{}}Ratio-based (Ratio of validation \\ accuracy and validation loss)\end{tabular} \\ \hline
\textbf{GA Operators} &
  \begin{tabular}[c]{@{}c@{}}Inheritance and hybrid \\ crossover, mutation\end{tabular} &
  \begin{tabular}[c]{@{}c@{}}Single point crossover \\ and mutation\end{tabular} &
  \begin{tabular}[c]{@{}c@{}}Single point crossover, and\\ shrink mutation\end{tabular} &
  \begin{tabular}[c]{@{}c@{}}Single point crossover \\ and mutation with ranking\\ selection\end{tabular} \\\hline
\textbf{\#Generations} &
  8 &
  10 &
  50 &
  40 \\\hline
\textbf{Population Size} &
  40 &
  50 &
  100 &
  30 \\\hline
\textbf{\begin{tabular}[c]{@{}c@{}}Child Model \\ and Dataset\end{tabular}} &
  \begin{tabular}[c]{@{}c@{}}ResNet38 (CIFAR10), \\ ResNet20 (CIFAR100 \\ \& Tiny ImageNet)\end{tabular} &
  \begin{tabular}[c]{@{}c@{}}WRN-28-10 (CIFAR10),\\ WRN-40-4 (CIFAR100)\end{tabular} &
  \begin{tabular}[c]{@{}c@{}}3 Different Custom \\ Networks (CIFAR10, \\ MNIST, \& Fashion \\ MNIST)\end{tabular} &
  Custom Network (CIFAR10) \\\hline
\textbf{\begin{tabular}[c]{@{}c@{}}Proposed Activation \\ Function\end{tabular}} &
  ELiSH and HardELiSH & \begin{tabular}[c]{@{}c@{}}
  $\sigma(x)\cdot erf(x)$, \\$tanh(x)\cdot min(x,0)$ \end{tabular}&
  \begin{tabular}[c]{@{}c@{}}Proposed 3 activation \\ functions for CIFAR10 \\ and 2 for MNIST dataset\end{tabular} &
  \begin{tabular}[c]{@{}c@{}}4 high-performing activation \\ function of EELU series\end{tabular} \\\hline
\textbf{\begin{tabular}[c]{@{}c@{}}Models Tested to\\ Study Generalization\end{tabular}} &
  ResNet56 \& VGG16 &
  \begin{tabular}[c]{@{}c@{}}None, but used the\\ same model for \\ higher epochs\end{tabular} &
  None &
  \begin{tabular}[c]{@{}c@{}}AlexNet, VGG16, ResNet50, \\ MobileNet, and Compact\\ Convolutional Transformer\end{tabular} \\\hline
\textbf{\begin{tabular}[c]{@{}c@{}}Datasets Tested to\\ Study Generalization\end{tabular}} &
  None &
  \begin{tabular}[c]{@{}c@{}}WRN-28-10 on CIFAR100,\\ WRN-40-4 on CIFAR10\end{tabular} &
  None &
  \begin{tabular}[c]{@{}c@{}}MNIST, Fashion MNIST,  Imagenette, \\ Beans,  Colorectal Histology, \\ CottonWeedID15, TinyImageNet\end{tabular} \\ \hline
\end{tabular}%
}}
\end{table*}
\section{Related Works}
{\color{myblue}NeuroEvolution (NE)} refers to the application of evolutionary algorithms in {\color{myblue}optimizing} neural network architectures and topology. Although the initial application of {\color{myblue}NE} was limited to {\color{myblue}evolving} the weights of a fixed architechture~\citep{morse2016simple}, recently{\color{myblue},} NE has been utilized to optimize the whole neural network and the network topology~\citep{elsken2019neural,wistuba2019survey}. \cite{real2017large} developed an evolutionary algorithm to discover the optimum network architecture through a novel intuitive mutation. The output of their approach is a fully trained model. In another work, \cite{real2019regularized} proposed a novel evolutionary approach by introducing an age property and obtained better classification accuracy on {\color{myblue}the} ImageNet dataset than other hand-designed architectures. \cite{hagg2017evolving} extended the {\color{myblue}NE} of Augmenting Topologies (NEAT) algorithm~\citep{stanley2002evolving} to evolve activation functions, weights{\color{myblue},} and topology of the network. Apart from {\color{myblue}NE,} some other approaches have also been proposed in this pursuit{\color{myblue},} such as Reinforcement Learning (RL)~\citep{ramachandran2017searching,baker2016designing} and Monte Carlo tree search~\citep{negrinho2017deeparchitect}.

In addition to optimizing neural network architectures, some studies have also focused on discovering improved activation functions. A summary of the related works can be found in {\color{myblue}Table}~\ref{tab:table-a}. \cite{basirat2018quest} proposed a Genetic Algorithm (GA) approach for the first time to search for activation functions, motivated by NEAT algorithm~\citep{stanley2002evolving}. However, they have limited their search space to the existing activation functions as candidate elements for generating new activation functions. {\color{myblue}For example, the candidate function list included ReLU, ELU, SeLU, Swish, Softplus, Sigmoid, etc. \cite{lapid2022evolution} employed similar search space but with Cartesian Genetic Programming (CGP). It is noteworthy that restricting the search space to only existing activation functions may result in suboptimal outcomes.} \cite{bingham2020evolutionary} adopted the search space used by \cite{ramachandran2017searching} and proposed a GA{\color{myblue}-}based activation function search algorithm for the ResNet20 {\color{myblue}architecture}. They concluded that it is feasible to evolve activation functions using smaller architecture and datasets {\color{myblue}and} then transfer them to larger models and difficult datasets within the same domain (e.g.{\color{myblue},} image classification). \cite{zahedinasab2020neuroevolutionary} developed a neural network with adaptive activation functions to investigate how this adaptation can improve image classification accuracy. \cite{nader2021evolution} {\color{myblue}have} utilized GA to develop activation function{\color{myblue}s} for different tasks such as multi-variate classification, regression{\color{myblue},} and image classification. {\color{myblue}Similar to} \cite{basirat2018quest}, they have also limited their search space to the existing activation functions as candidate elements for generating new activation functions~\citep{nader2021evolution}. \cite{marchisio2018methodology} introduced a methodology to improve overall Deep Neural Network (DNN) performance by using {\color{myblue}a} layer-wise activation function. In another study, \cite{singh2020pursuit} proposed a genetic algorithm and used basic mathematical functions and existing activation functions as candidate elements.

\cite{ramachandran2017searching} proposed another activation function development framework using {\color{myblue}Reinforcement} Learning (RL) and concluded that carefully selecting the candidate elements of the activation function can result in developing high-performing activation functions. For example, a common structure shared by top activation functions generated by them is $b(x, u(x))$, where $b$ \& $u$ are binary and unary functions{\color{myblue},}  respectively. Besides choosing the search space, defining an effective fitness function for the GA is also crucial. \cite{nader2021evolution} and  \cite{ramachandran2017searching} used validation accuracy as a fitness function and reward function, respectively, while \cite{bingham2020evolutionary} used both of the validation loss and validation accuracy separately. {\color{myblue}\cite{bingham2020evolutionary} concluded that the accuracy-based fitness function favors exploration over exploitation. Here exploration refers to the fact that activation functions with poor validation accuracy may still have a reasonable chance of surviving to the next generation. For instance, an activation function that achieves 90\% validation accuracy is only 2.2 times more likely to be selected for the next generation than a function with only 10\% validation accuracy since $e^{0.9}/e^{0.1}\approx 2.2$ {\color{myblue}~\citep{bingham2020evolutionary}}. On the other hand, loss-based fitness functions favor exploitation by penalizing the poor activation functions and encouraging the current best ones to move forward to the next generation. For instance, an activation
function with a validation loss of 10 is 21,807 times less likely to
be chosen for the next generation compared to a function with a
validation loss of 0.01 because $e^{-0.01}/e^{-10}\approx 21807$ \citep{bingham2020evolutionary}}. 

A noticeable similarity among studies mentioned in the literature is the use of tree{\color{myblue}-}based search space that was first introduced by ~\cite{ramachandran2017searching}. {\color{myblue}Among the GA operators, crossover and mutation appeared to be prevalent. While crossover inspires evolution to discover better activation functions more quickly than random search, mutation enables evolution to explore the search space \citep{bingham2020evolutionary}. Moreover, mutation helps prevent better-performing activation functions from skewing the early generations of the activation function search \citep{bingham2020evolutionary}.} \cite{ramachandran2017searching} concluded that less intricate activation functions generally outperform the intricate ones. The intricacy of the developed activation function can be defined as the number of core units needed to construct the activation function. A core unit can be formed as $b(u_1(x),u_2(x))$ where $u_1$ and $u_2$ represent two unary functions. \cite{ramachandran2017searching} also observed that the best-performing activation mostly has 1 or 2 core units.

{\color{myblue}In light of the literature, our study contributes to the following points:


\begin{enumerate}
\item \textbf{Improved evolutionary search:}
    \begin{itemize}
        \item \textit{Enhanced search space} - Instead of using only the existing activation functions as candidate elements for the evolutionary algorithm as done in previous works, we have incorporated a comprehensive list of candidate functions (See Section 3.1.1), resulting in a search space with more than 170 million combinations of candidate solutions. We have added several modified binary functions to the search space e.g., $x_1\cdot erf(x_2)$, which eventually helped in finding high-performing activation functions.
        \item \textit{Efficient fitness function} - We proposed a ratio-based fitness function by taking the ratio of validation accuracy and validation loss due to its high likelihood of selecting better chromosomes for the next generation.
    \end{itemize}
\item \textbf{Novel activation functions:} A series of high-performing activation functions called EELU for image recognition tasks have been proposed.
\item \textbf{Study of generalization:} We studied the generalization of the proposed activation functions across eight image recognition datasets and five state-of-the-art neural network architectures.
\end{enumerate}
}

\section{Methodology}
\label{sec:experiment}
{\color{myblue}This} paper proposes an Activation Function Optimization Scheme (AFOS) for image classification tasks in deep neural networks. The following sections include the specific details of the implementation of the optimization scheme. First, we have introduced the notations used to describe the AFOS.

\nomenclature{$\Phi$}{Base neural network}
\nomenclature{$P_0$}{Initial population}
\nomenclature{$\Omega$}{Search space}
\nomenclature{$F$}{Fitness function}
\nomenclature{$C$}{Chromosome}
\nomenclature{$f$}{Fitness score}
\nomenclature{$\Theta$}{Termination criterion}
\nomenclature{$P$}{Population}
\nomenclature{$P_{next}$}{Population of the next generation}
\nomenclature{$M$}{Mutation}
\nomenclature{$n_m$}{Number of chromosomes generated by mutation}
\nomenclature{$X$}{Crossover}
\nomenclature{$n_x$}{Number of chromosomes generated by crossover}
\nomenclature{$\Psi$}{Mating pool}
\nomenclature{$S$}{Selection}
\nomenclature{$n_{\psi}$}{Number of chromosomes generated by selection}
\nomenclature{$J$}{Number of chromosomes in each population}
\nomenclature{$K$}{Number of times Genetics Algorithm is repeated}
\nomenclature{$D_{Tr}$}{Training dataset}
\nomenclature{$D_{V}$}{Validation dataset}
\nomenclature{$D_{Te}$}{Test dataset}
\nomenclature{$AF$}{Activation function}
\nomenclature{$V_a$}{Validation accuracy}
\nomenclature{$V_l$}{Validation loss}
\nomenclature{$N$}{Number of generations}
\nomenclature{$\rho$}{Number of classes in the datasets}
\nomenclature{$\tau$}{Cross site for crossover}
\nomenclature{$x$}{Probability of crossover}
\nomenclature{$m$}{Probability of mutation}
\nomenclature{$n_{rand}$}{Number of chromosomes generated by random initialization}
{\color{myblue}
\printnomenclature
}
\begin{figure*}[htp]
  \centering
    \includegraphics[width=\linewidth]{ 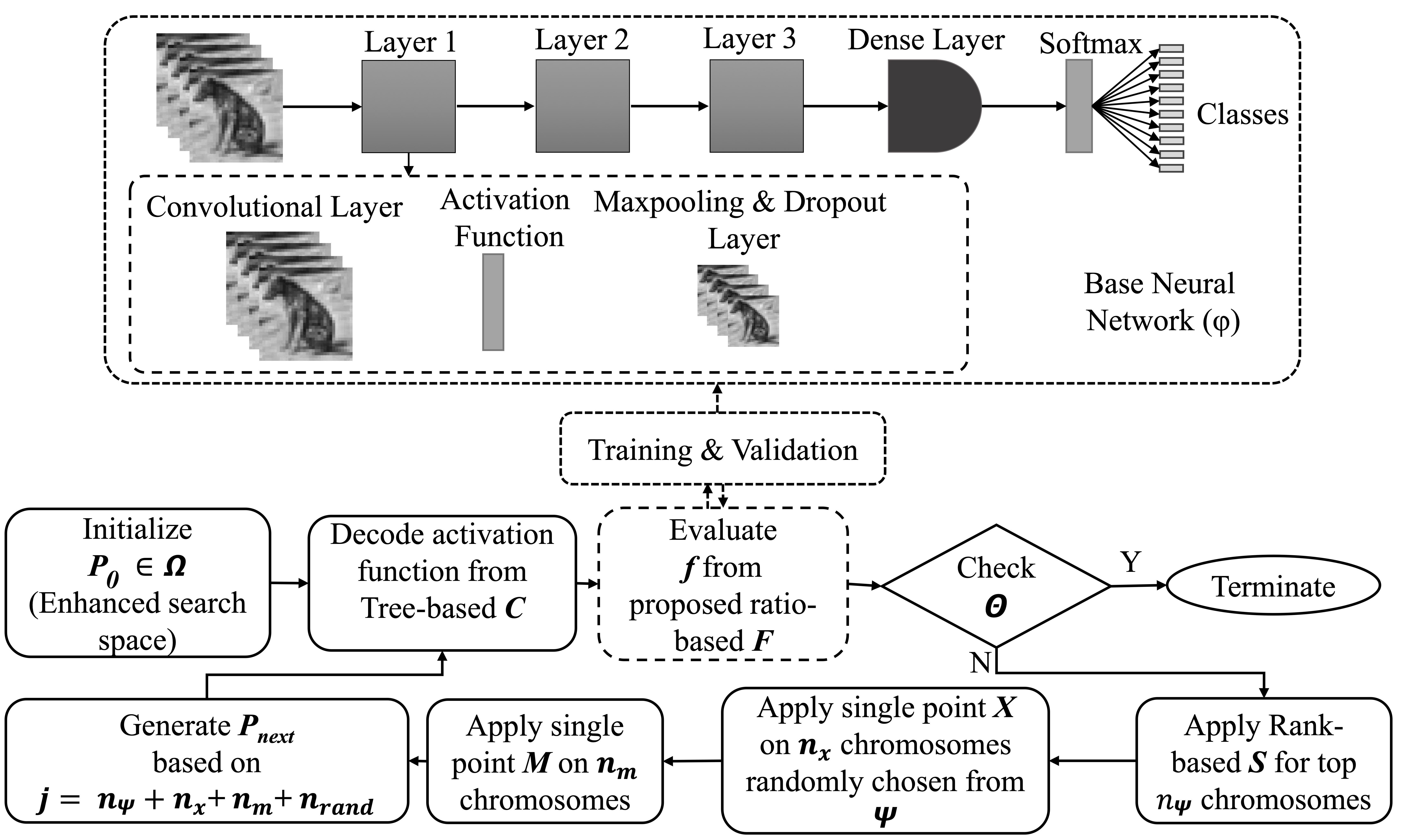}
  \caption{Activation Function Optimization Scheme (AFOS)}
  \label{fig:fig-1}
\end{figure*}
\subsection{Activation Function Optimization Scheme (AFOS)}
{\color{myblue}This work employed} the Genetic Algorithm (GA) as the optimization method. {\color{myblue}GA is a meta-heuristic and a subclass of evolutionary algorithms inspired by natural selection.} Although GA, as a meta-heuristic, does not guarantee to find the optimal solution, it evolves and eventually generates a near-optimal solution. We first developed a chromosome representation of the activation function for the population $P$ in GA. Then, the initial population $P_{0}$ was generated and evolved toward better solutions. Next, a fitness function $F$ was evaluated to find the fitness score $f$ for $P_{0}$. Then the algorithm checks the termination criteria $\Theta$ and terminates if the criteria {\color{myblue}are fulfilled}. Otherwise, the algorithm applies the selection $S$, crossover $X$, and mutation $M$ to generate population for the next generation $(P_{next})$. {\color{myblue}The number} of chromosome{\color{myblue}s} in each population is $J$. We repeated the GA for $K$ times. Figure \ref{fig:fig-1} and Algorithm \ref{alg-1} illustrate the framework of AFOS.

\begin{figure}[h]
  \centering
  \includegraphics[width=0.5\linewidth]{ 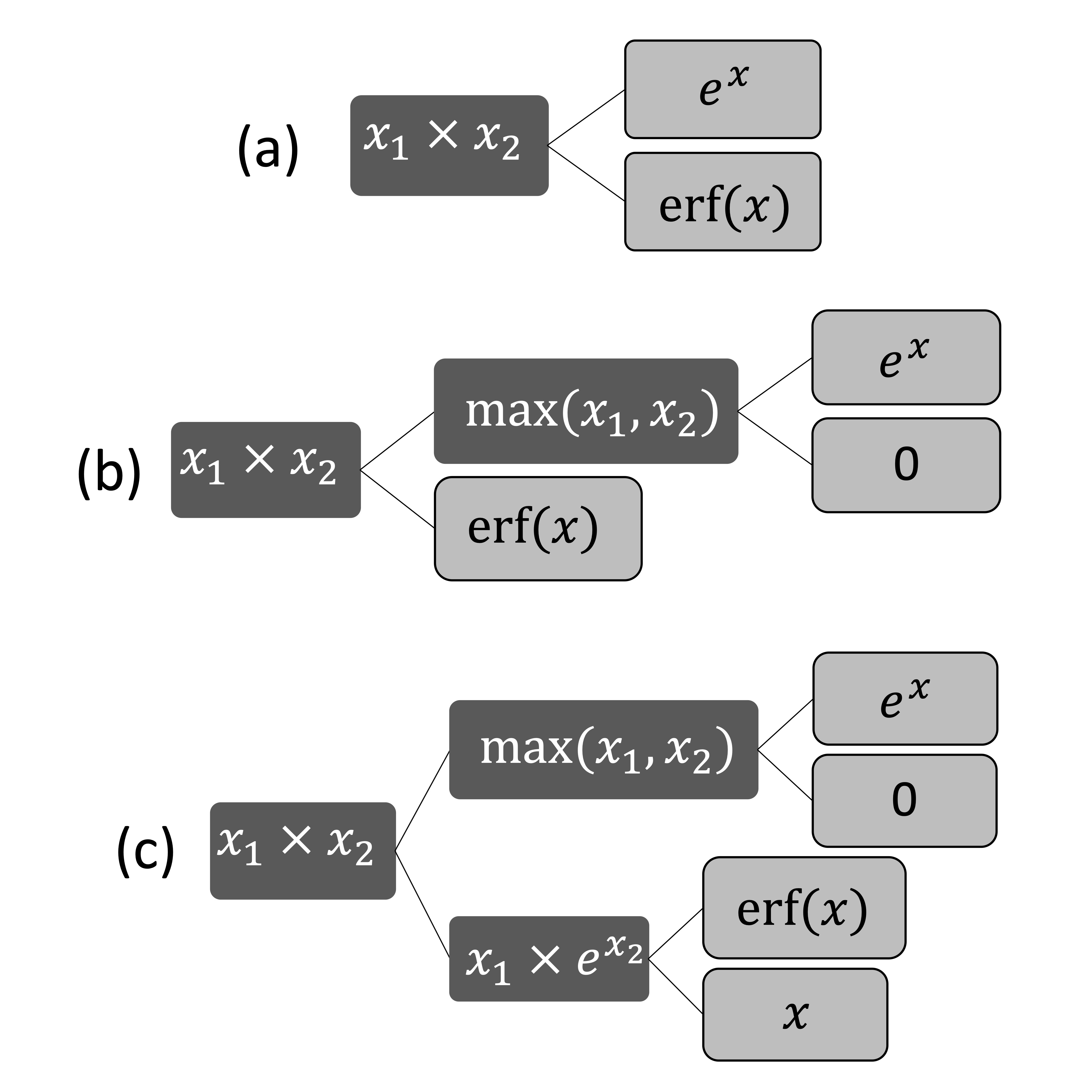} 
   \caption{Tree-based chromosome representation. The unary functions are depicted as light gray color and binary functions are represented as dark gray color. The activation function represented here are: (a) $e^x\cdot erf(x)$, (b) $max(e^x,0)\cdot erf(x)$, and (c) $max(e^x,0)\cdot erf(x)\cdot e^x$}
   \label{fig:fig-2}
\end{figure}
\subsubsection{Chromosome Representation $(C)$}
In this study{\color{myblue},} we used the tree{\color{myblue}-}based encoding similar to the encoding used by \cite{ramachandran2017searching}. Tree{\color{myblue}-}based encoding is generally suitable for evolving expressions or functions~\citep{kumar2013encoding}. We represented each activation function as a tree consisting of unary $(u)$ and binary functions $(b)$. While $u$ takes {\color{myblue}a} single input, $b$ takes two inputs as illustrated in Figure \ref{fig:fig-2}. We defined our enhanced search space ($\Omega$) by using the following list of unary and binary functions:
\begin{itemize}
    \item Unary functions: $x$, $-x$, $e^x$, $|x|$, $e^{-x}$, $min(x,0)$, $max(x,0)$, $sin(x)$, $cos(x)$, $sinh(x)$, $tanh(x)$, $\sin^{-1}(x)$, $\tan^{-1}(x)$, $\arcsinh(x)$, $\arctanh(x)$, $erf(x)$, $\sigma(x)$, $xerf(x)$, $\ln (1+e^x)$
    \item Binary functions: $x_1+x_2$, $x_1\cdot x_2$, $max(x_1,x_2)$, $min(x_1,x_2)$, $max(max(x_1,0), x_2)$, $max(min(x_1,0), x_2)$,  $min(max(x_1,0), x_2)$, $min(min(x_1,0), x_2)$, $x_1\cdot e^{x_2}$, $x_1\sigma (x_2)$, $x_1\cdot erf(x_2)$
\end{itemize}
where, $\sigma (x) = \frac{1}{1+e^{-x}}$ is the sigmoid function and $erf(x) = \frac{2}{\sqrt{\pi }}\int_{0}^{x}e^{-t^2}dt$ is the gaussian error function. Although we constrained the maximum number of core units to 2, this search space had {\color{myblue}more than} 170 million combinations of candidate solutions. 

\begin{table}[h]
\centering
\caption{CNN architecture $(\phi)$ used in this study. $\rho$ is the number of classes for any dataset.}
\label{tab:phi}
\resizebox{0.5\linewidth}{!}{
\begin{tabular}{@{}llll@{}}
\hline
Layers     & Filter-size & Kernel-size & Output       \\ \hline
Conv     & 28          & 3×3         & (32, 32, 28) \\
Conv     & 32          & 3×3         & (32, 32, 32) \\
Max-pool & -           & 2×2         & (16, 16, 32) \\
Dropout  & -           & -           & (16, 16, 32) \\
Conv     & 64          & 3×3         & (16, 16, 64) \\
Conv     & 64          & 3×3         & (16, 16, 64) \\
Max-pool & -           & 2×2         & (8, 8, 64)   \\
Dropout  & -           & -           & (8, 8, 64)   \\
Conv     & 128         & 3×3         & (8, 8, 128)  \\
Conv     & 128         & 3×3         & (8, 8, 128)  \\
Max-pool & -           & 2×2         & (4, 4, 128)  \\
Dropout  & -           & -           & (4, 4, 128)  \\
Flatten  & -           & -           & (1, 1, 2048) \\
Dense    & $\rho$          & -           & (1, 1, $\rho$)   \\ \hline
\end{tabular}}
\end{table}

\begin{algorithm}
{\color{myblue}
	\caption{Activation Function Optimization Scheme (AFOS)}
	\label{alg-1}
	\begin{algorithmic}[1]
		\State \textbf{Input} $D_{Tr}$, $D_V$
		\State \textbf{Output} $AF$
		\For{$k$:=1 to K} 
		    \State Initialize $P_{0}$
		    \Repeat
		        \For{$j$:=1 to J}
		            \State Decode $C_j$
		            \State Train $\phi$ with $C_j$ on $D_{Tr}$
		            \State Evaluate $V_a$ and $V_l$ on $D_V$
		            \State $f \leftarrow  \frac{V_a}{V_l}$
		            \State $j \leftarrow j + 1$
		        \EndFor
		            \State Apply $S$, $X$, $M$
		            \State Generate $P_{next}$
		    \Until $\Theta$ is $True$ ($N>40$)
        \EndFor
	\end{algorithmic}
     }
\end{algorithm}

\subsubsection{Initial Population $(P_{0})$}
The initial population represents the individual candidate solutions for the problem. {\color{myblue}The population size depends mainly on the nature of the problem. For activation function development, previous studies have used a population size ranging from 30 to 100. Moreover, \cite{chen2012large} concluded that a large population size is not always helpful for evolutionary algorithms under several conditions. Therefore, in this study, we set the population size $J$ to 30 following the work of \citep{li2022gaaf} and initiated the population randomly.} The advantage of a random initial population is that it can potentially initiate values from the entire range of the search space.

\subsubsection{Fitness Function $(F)$}
To calculate the fitness score $f$, the algorithm first decode{\color{myblue}s} the chromosome $C$. For each chromosome $C$, the algorithm replaced the ReLU activation function in the base neural network $\phi$ as shown in Table~\ref{tab:phi} with the decoded activation function. The base neural network $\phi$ was trained with {\color{myblue}the} CIFAR10 dataset for 15 epochs. As the CIFAR10 dataset does not have a separate validation set, we split the training set into two sets: training $(D_{Tr})$ and validation $(D_V)$ sets following the work of \cite{bingham2020evolutionary}. The test set $(D_{Te})$ was kept unchanged and used in the next stage of the experiment. The algorithm then evaluated the trained model on the validation dataset and calculated the value of the fitness score $f$, where $f = V_a/V_l$. We chose this ratio{\color{myblue}-}based fitness function due to its high likelihood of selecting better chromosome{\color{myblue}s} for the next generation. For instance, a hypothetical activation function with a validation accuracy of 0.9 and validation loss of 0.01 will be $1.2\times 10^{39}$ times more likely to be chosen for the next generation {\color{myblue}than} an activation function with 0.1 validation accuracy and 10 validation loss. $e^\frac{0.9}{0.01}/e^\frac{0.1}{10}\approx 1.2\times 10^{39}$. 

\subsubsection{Selection $(S)$}
The process of choosing two parents from the population for generating offspring is referred {\color{myblue}to} as selection. The notion of selection is to prioritize fitter individuals so that their offspring can have higher fitness. Roulette, tournament{\color{myblue},} and rank selection are the {\color{myblue}most} popularly used selection methods. In our study{\color{myblue},} we focused on ranking the chromosomes according to their fitness score $f$ and selecting the {\color{myblue} ones with} higher scores. Through the selection operator{\color{myblue},} a mating pool $\Psi$ consisting of $n_{\Psi}$ number of chromosomes was generated for crossover $X$.
\begin{figure}[h]
  \centering
  \includegraphics[width=0.5\linewidth]{ 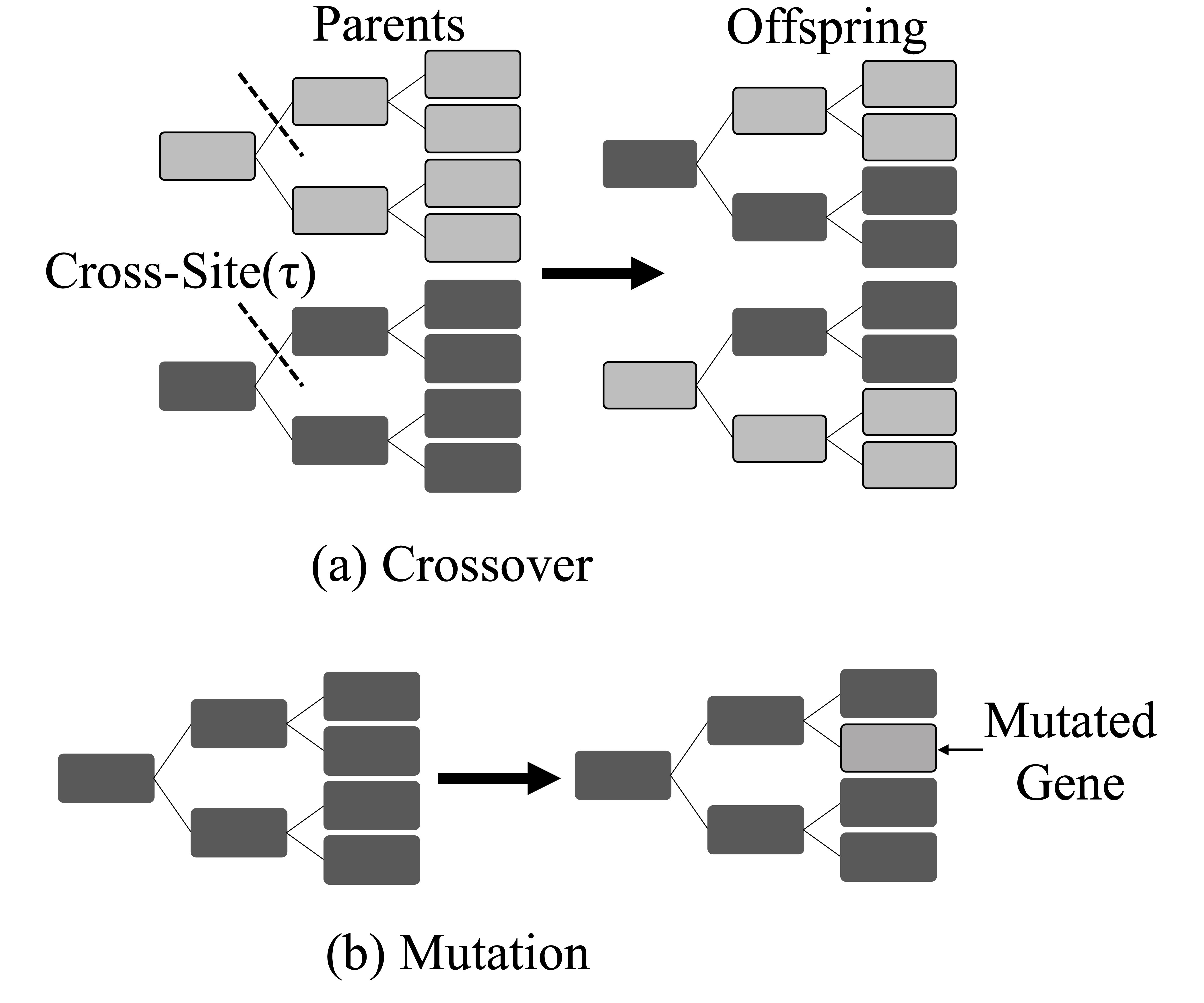} 
   \caption{(a) Generating offspring through crossover of two parent chromosomes and (b) Randomly replaced gene (gray colored) through mutation}
   \label{fig:fig-3}
\end{figure}
\subsubsection{Crossover $(X)$}
{\color{myblue}
Crossover involves taking two chromosomes as parents and generating offspring from them. The crossover process typically includes three steps: first, selecting two random chromosomes from the mating pool \( \Psi \); second, choosing a random crossover point \( \tau \); and third, swapping the position values along the chromosome string based on a crossover probability \( x \). Figure \ref{fig:fig-3}(a) illustrates an example of this crossover process. In our study, new \( n_x \) chromosomes were produced using a single-point crossover method.}

\subsubsection{Mutation $(M)$}
{\color{myblue}
Mutation involves the random alteration of a gene within a chromosome. While crossover refines existing solutions to discover superior ones, mutation serves as a mechanism to explore the entire search space. This process is crucial for preventing the algorithm from becoming stuck in local optima. In our algorithm, we randomly selected a chromosome from the entire population and applied a single-point mutation with a probability of \( m \). Figure \ref{fig:fig-3}(b) illustrates a single-point mutation, where a dark gray block (gene) is replaced with a light gray block (gene) at random. To maintain the consistency of the activation function, the algorithm replaced unary functions with other unary functions and binary functions with other binary functions. The number of chromosomes produced through mutation is \( n_m \).
}

\subsubsection{Population for Next Generation $(P_{next})$}
{\color{myblue}
From the crossover of the \( n_{\Psi} \) highest-ranked chromosomes, we generated \( n_x \) new evolved chromosomes. Additionally, we applied single-point mutation to \( n_m \) chromosomes. Given that the population size for each generation is \( J \), the remaining \( n_{\text{rand}} \) chromosomes were created randomly. Thus, \( n_{\text{rand}} = J - (n_{\Psi} + n_x + n_m) \), which induced the population diversity.
 }
\section{Experiments and Results}
\label{sec:experiments}
In this section, we describe the experimental setup and results from the AFOS and investigate the robustness of the results by analyzing the properties of the evolved activation functions. We conducted our experiments using GPU NVIDIA GeForce RTX 2080 SUPER on Intel Core i9-10900, 32.0 GB RAM, and 64-bit Windows 10 Operating System. We have made the code available at: \url{https://github.com/abdurrahman1828/AFOS}.

\begin{figure}[h]
  \centering
   \includegraphics[width=0.45\linewidth]{ 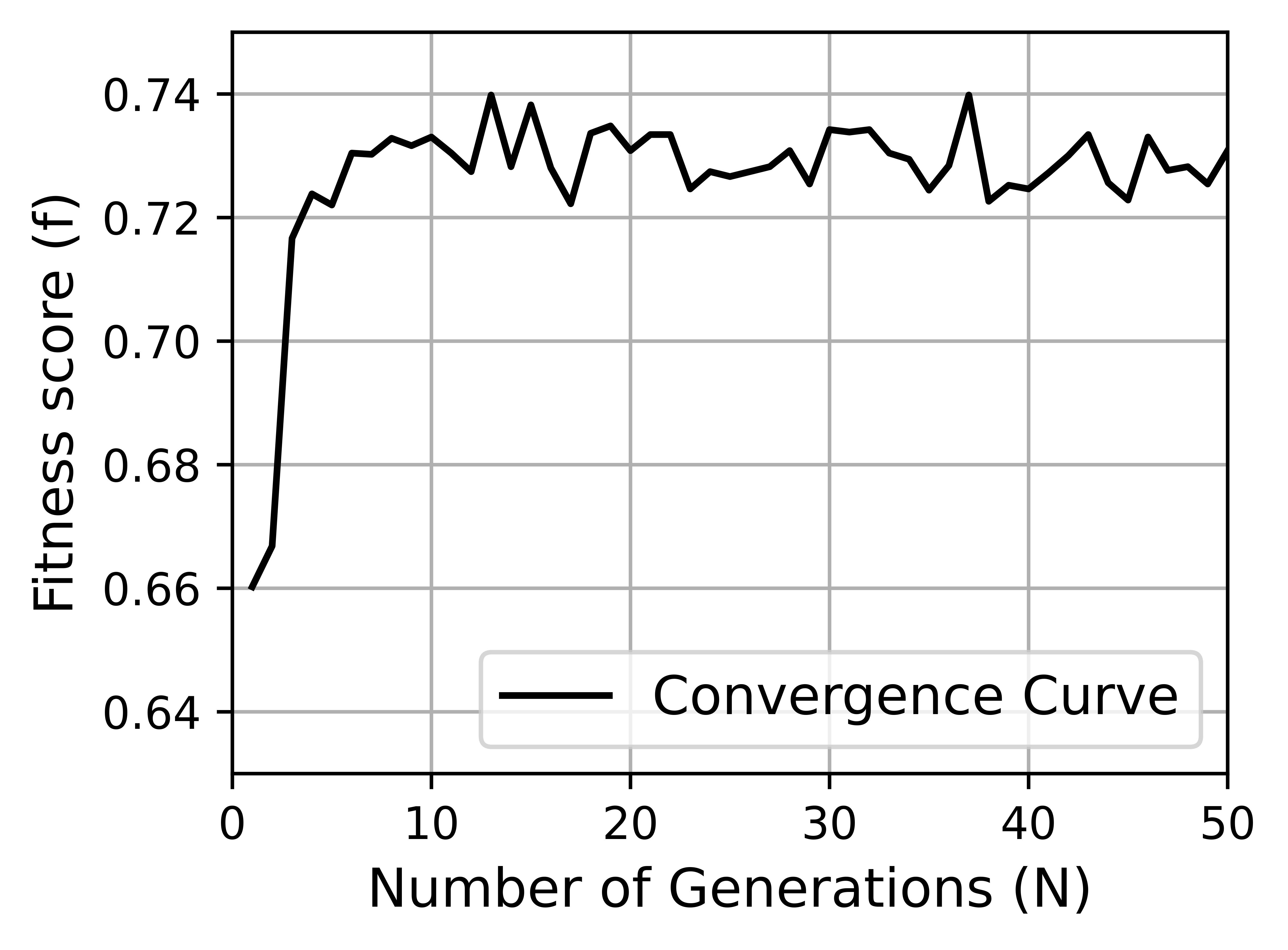}
   \caption{Convergence curve of AFOS.}
   \label{fig:fig-4}
\end{figure}
\subsection{Experimental Setup}
{\color{myblue}
In the AFOS, we utilized a base neural network \( \phi \) as outlined in Table~\ref{tab:phi} for calculating the fitness score. This neural network features a typical deep learning architecture, consisting of consecutive convolutional, max-pooling, and dropout layers, followed by flatten and dense layers in the final stages. The purpose of selecting this base network was to evaluate how the activation functions derived from it perform across different neural architectures. Following the prevalent trends in the literature, we selected the CIFAR-10 dataset \citep{Krizhevsky09learningmultiple} as the primary dataset for running experiments on AFOS. Each chromosome was trained for 15 epochs with a batch size of 64. The optimizer used was Stochastic Gradient Descent (SGD) with a learning rate of 0.01 and momentum set at 0.9. We employed the \verb|he_uniform| initializer for the kernel and maintained the padding option as \verb|same|. The loss function utilized was \verb|categorical_crossentropy|. The detailed parameters for the optimization algorithm are provided in \ref{sec:appendix_c}. As illustrated in the convergence graph in Figure \ref{fig:fig-4}, significant improvements were not observed beyond 40 generations. Hence, we set the termination criterion \( \Theta \) for AFOS to be 40 generations. The AFOS method required 1 to 1.5 GPU days under varying initial conditions. To optimize runtime, we implemented two stopping criteria: first, halting evaluation if the loss function value became undefined, and second, stopping if the accuracy after the first epoch did not exceed a threshold of 0.25. This allowed us to proceed efficiently to the next chromosome.

For different initial conditions, we initialized AFOS with various random seeds to create the initial populations for five replications. This approach revealed a consistent prevalence of the $erf(x)$ unary function and the $x \cdot f(x)$ binary function among the top-performing activation functions in these replications.} {\color{myblue}We selected the top 15 activation functions from AFOS based on the validation accuracy of the model on the CIFAR10 dataset.} Then, we trained the base neural network $\phi$ with the developed activation functions from scratch for 50 epochs and evaluated them on the test set $(D_{Te})$ of CIFAR10. In Table~\ref{tab:tab-y}, we mentioned the expressions of the developed activation functions and their test accuracy (mean and standard deviation) of 5 replications.
\begin{table}[h]
\centering
\caption{Accuracy of the developed activation functions with 50 epochs and 5 replications}
\label{tab:tab-y}
\resizebox{0.5\linewidth}{!}{
\begin{tabular}{@{}cc@{}}
\hline
Activation Function          & Mean ± St. Dev       \\ \hline
$erf(x)\cdot ln(1+e^x)$                      & 82.74 ± 0.14 \\\hline
$-x\cdot erf(e^{-x})$                       & 82.67 ± 0.30 \\\hline
$x\cdot erf(e^x)$                      & 82.66 ± 0.29 \\\hline
$x\cdot (erf(e^x))^2$                     & 82.48 ± 0.61 \\\hline
$max(x,min(x,0))\cdot erf(min(x,0)))$         & 82.46 ± 0.54 \\\hline
$max(x,0)\cdot erf(\tan^{-1}(x))-x$         & 82.44 ± 0.37 \\\hline
$ln(1+e^x)\cdot erf(erf(x))$               & 82.40 ± 0.47 \\\hline
\begin{tabular}[c]{@{}c@{}}$(\tanh (x)+max(x,0))\cdot$\\$\sigma (ln(1+e^x)\cdot erf(\tan^{-1}(x)))$\end{tabular} & 82.38 ± 0.39 \\\hline
$ln(1+e^x)\cdot \tanh (x)$              & 82.29 ± 0.18 \\\hline
$max(sin(x)\sigma (sin(x)), (x\cdot erf(x)+\tanh (x)))$ & 82.27 ± 0.35 \\\hline
\begin{tabular}[c]{@{}c@{}}$(\tanh (x)+max(x,0))\cdot$\\$\sigma (ln(1+e^x)\cdot erf(\sinh (x)))$\end{tabular} & 82.26 ± 0.13 \\\hline
$x\cdot erf(x)+sin(x)$               & 82.25 ± 0.47 \\\hline
$max(x, erf(x))$             & 82.19 ± 0.51 \\\hline
$max(x,0)\cdot \tan^{-1}(x)-x$         & 82.15 ± 0.29 \\ \hline
$max(ln(1+e^x)\cdot \tanh (x),erf(x))$               & 82.15 ± 0.33 \\\hline
\end{tabular}%
}
\end{table}
\begin{table*}[h]
\centering
\caption{Mathematical expression and implementation of the top four developed activation functions}
\label{tab-m}
\resizebox{0.8\linewidth}{!}{
\begin{tabular}{@{}ccc@{}}
\hline
Name   & Mathematical Expression & TensorFlow Implementation \\ \hline
EELU-1 &$x\cdot erf(e^x)$ & \texttt{x*tf.math.erf(tf.math.exp(x))} \\ \hline
EELU-2 & $-x\cdot erf(e^{-x})$ & \texttt{x*tf.math.erf(tf.math.exp(x))} \\ \hline
EELU-3 & $x\cdot (erf(e^x))^2$ & \texttt{x*tf.math.pow(tf.math.erf(tf.math.exp(x)),2)} \\ \hline
EELU-4 & $erf(x)\cdot ln(1+e^x)$& \texttt{tf.math.erf(x)*tf.math.log(1+tf.math.exp(x))}                      \\ \hline
\end{tabular}}
\end{table*}
\begin{figure*}[h]

  \centering
  \subfloat[]{\includegraphics[width=0.3\linewidth]{ 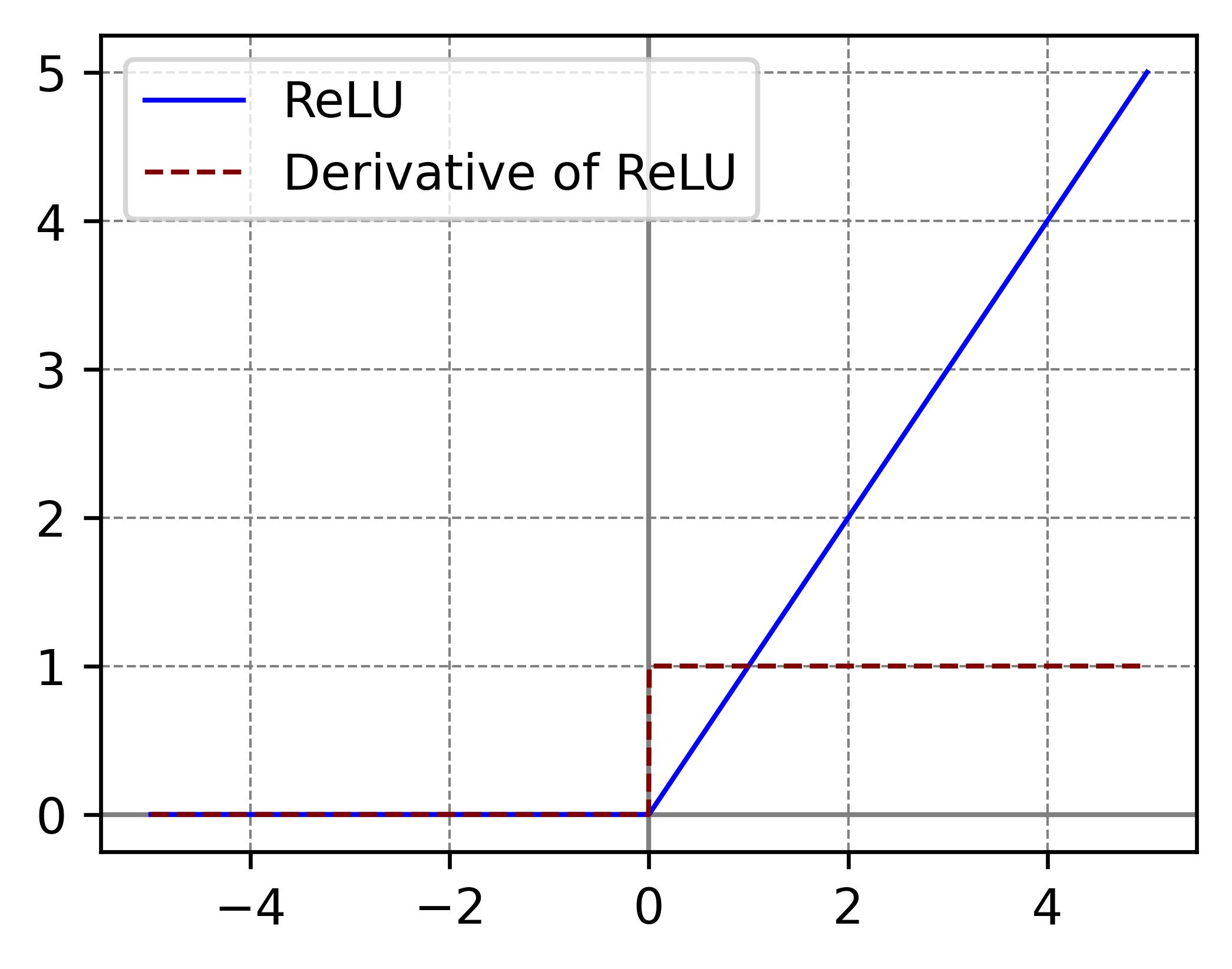}  
  \label{fig:EELU-1}}
  \subfloat[]{\includegraphics[width=0.3\linewidth]{ 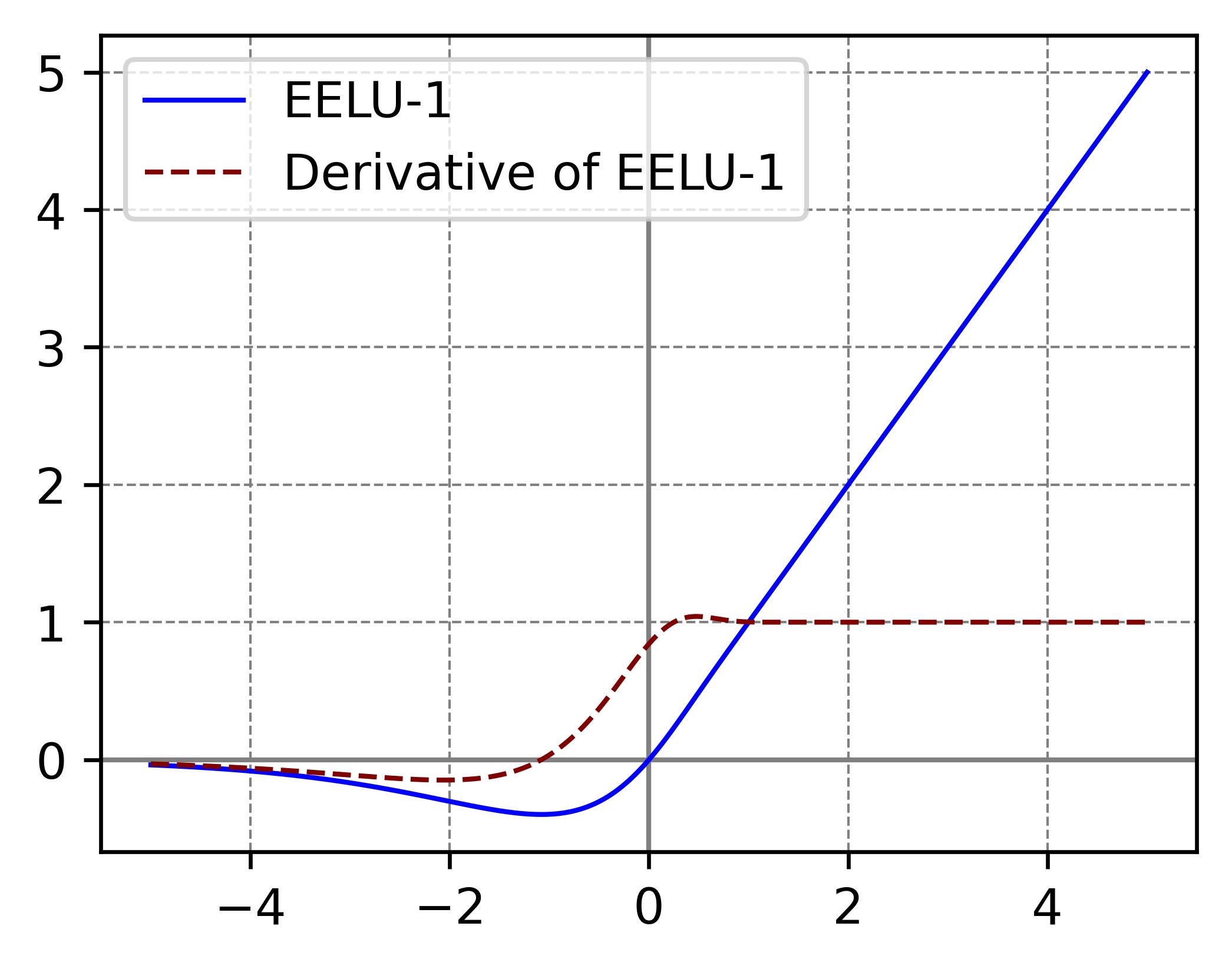}  
  \label{fig:EELU-1}}
  \subfloat[]{\includegraphics[width=0.3\linewidth]{ 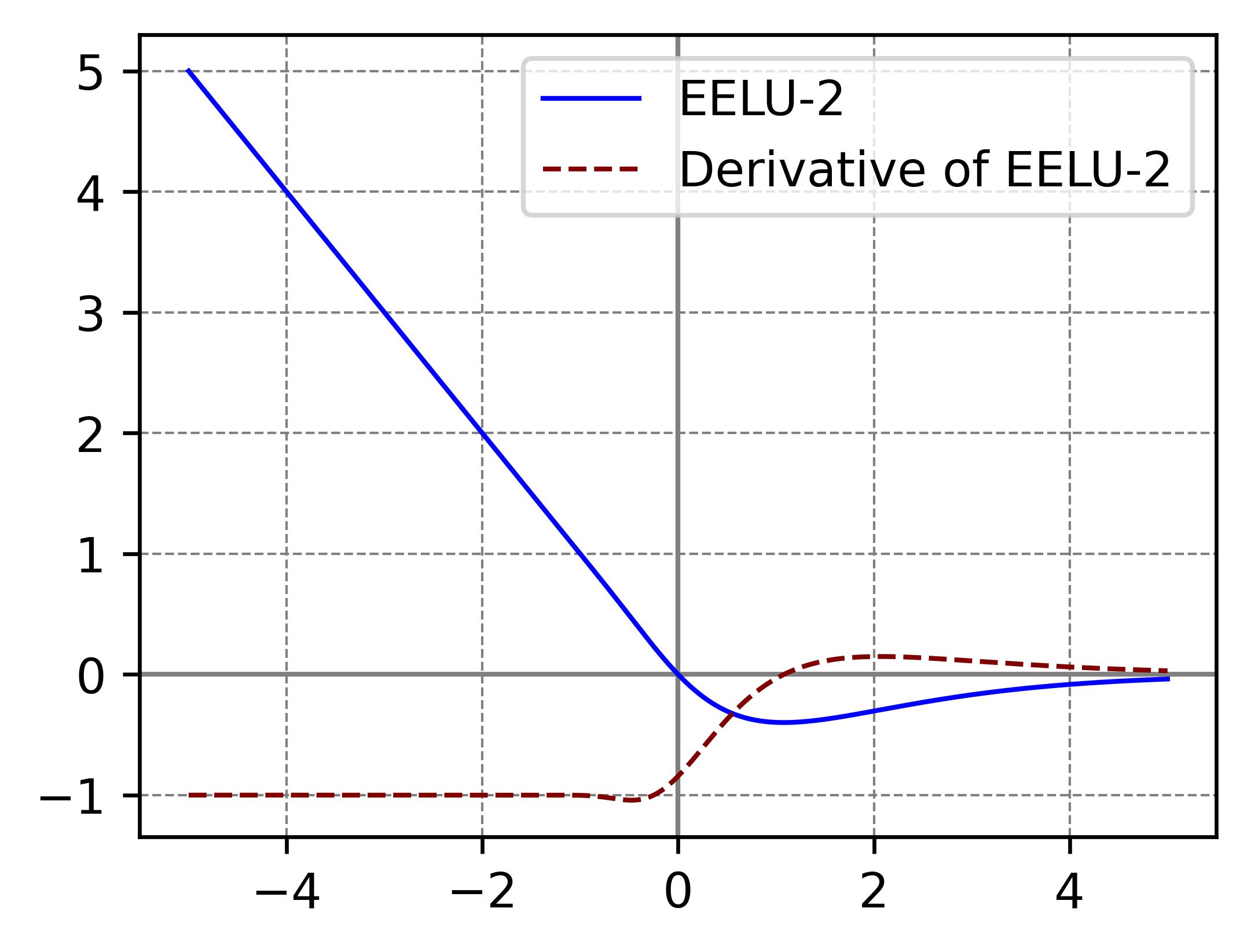}  
  \label{fig:EELU-1}}
\hfill
  \subfloat[]{\includegraphics[width=0.3\linewidth]{ 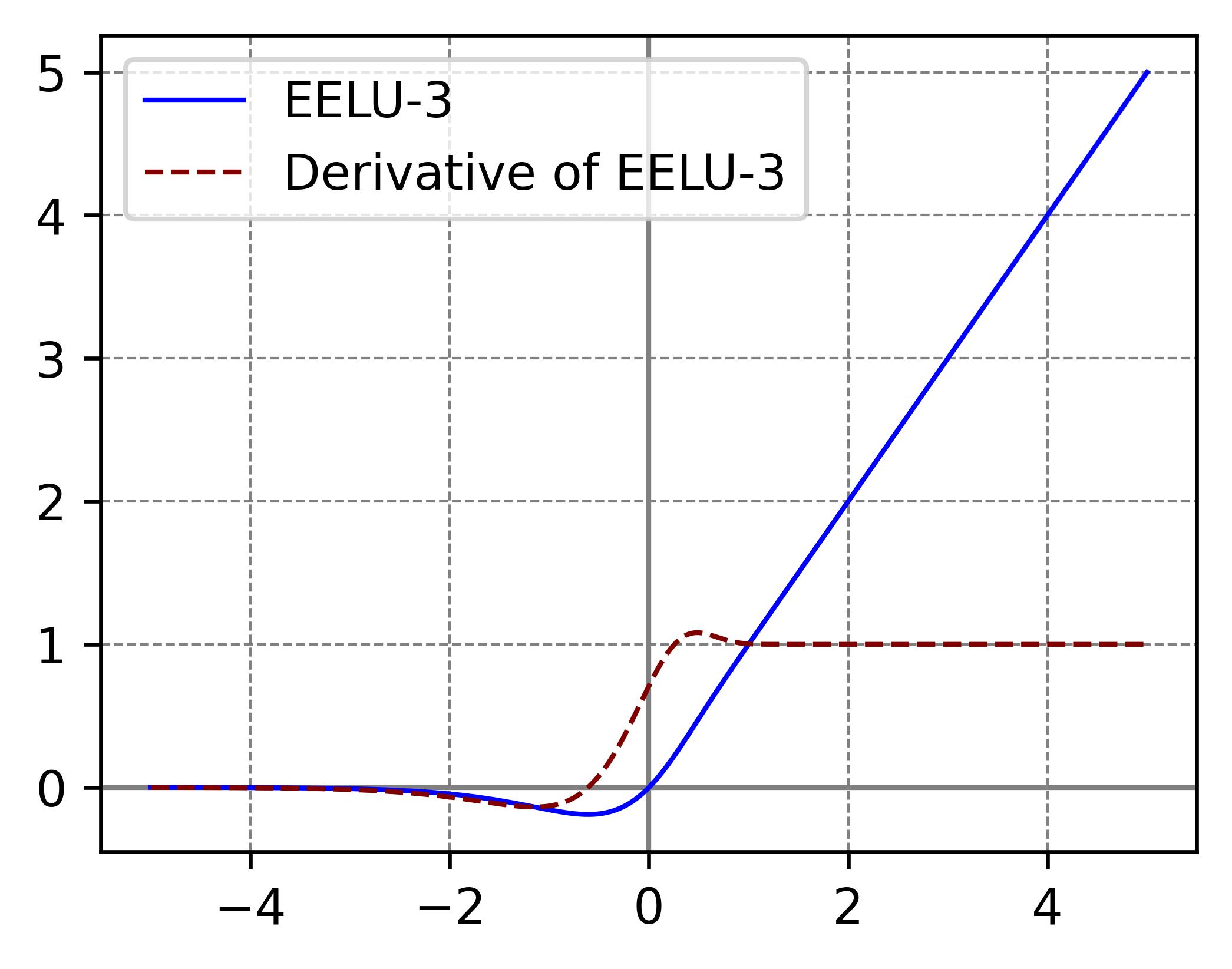}  
  \label{fig:EELU-1}}
  \subfloat[]{\includegraphics[width=0.3\linewidth]{ 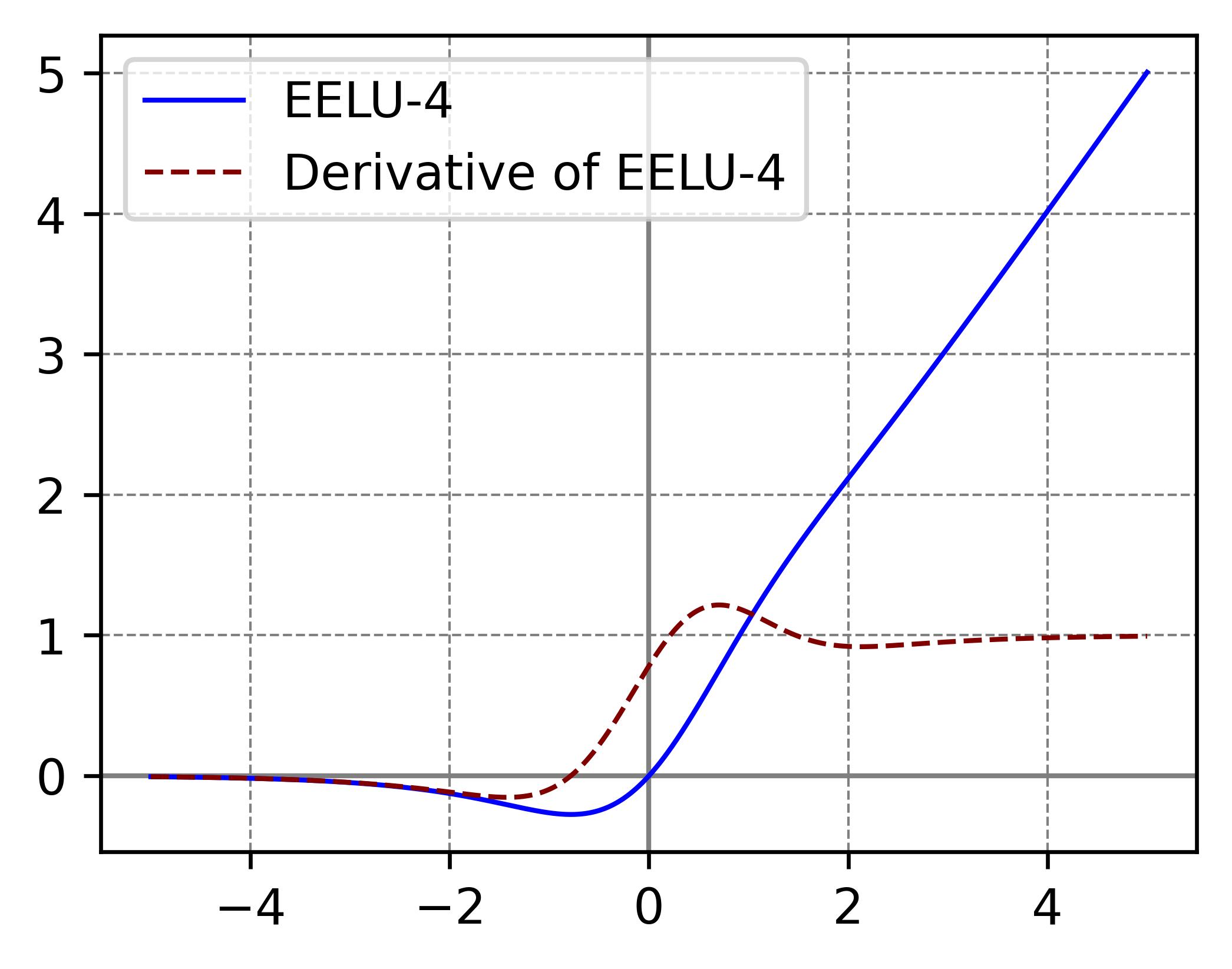}  
  \label{fig:EELU-1}}
\caption{(Best viewed in color) Graphical representation of ReLU and the developed activation functions. (a) ReLU, (b) EELU-1, (c) EELU-2, (d) EELU-3, (e) EELU-4}
\label{fig:fig-5}
\end{figure*}

{\color{myblue}
In the next step, we selected the top four developed activation functions, examined their properties, and performed comparative tests. The rationale for choosing the top four instead of three or five is that these four functions exhibit nearly similar and simpler mathematical expressions compared to other developed functions. All four of these activation functions incorporate the exponential and error functions, with three of them also including a linear \( x \) component. We refer to this group as the \textbf{Exponential Error Linear Unit (EELU)} series. Table~\ref{tab-m} presents the mathematical expressions and TensorFlow implementations of the EELU series. Notably, the order of these functions has been rearranged in Table~\ref{tab-m} according to their forms and structures.}

\begin{table}[h]
 \centering
 \caption{Comparison of properties of ReLU and EELU series}
\label{tab:table-properties}
\resizebox{0.6\linewidth}{!}{
\begin{tabular}{@{}lccccc@{}}
\hline
Property       & ReLU & EELU-1 & EELU-2 & EELU-3 & EELU-4 \\ \hline
Non-linear     &\cmark& \cmark&\cmark&\cmark&\cmark\\
Upper-bounded  &\xmark&\xmark&\xmark&\xmark&\xmark\\
Lower-bounded  &\cmark& \cmark&\cmark&\cmark&\cmark\\
Differentiable &\xmark&\cmark&\cmark&\cmark&\cmark\\
Non-monotonic &\xmark& \cmark&\cmark&\cmark&\cmark\\ \hline
\end{tabular}%
}
\end{table}
\begin{figure*}[h]
    \centering
    \subfloat[]{\includegraphics[width=0.3\linewidth]{ 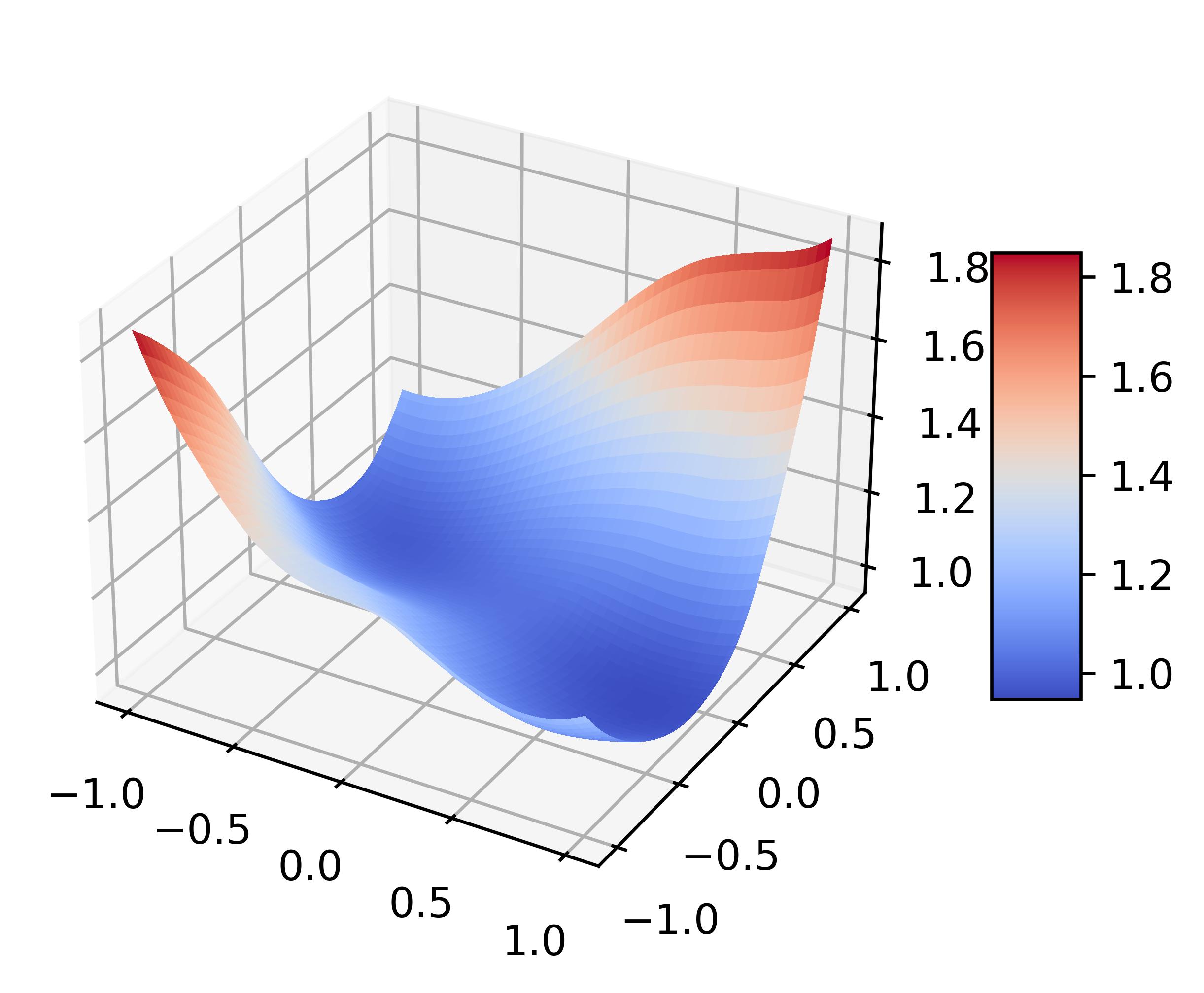}
    \label{fig:sub-first}}
    \hfill
    \subfloat[]{\includegraphics[width=0.3\linewidth]{ 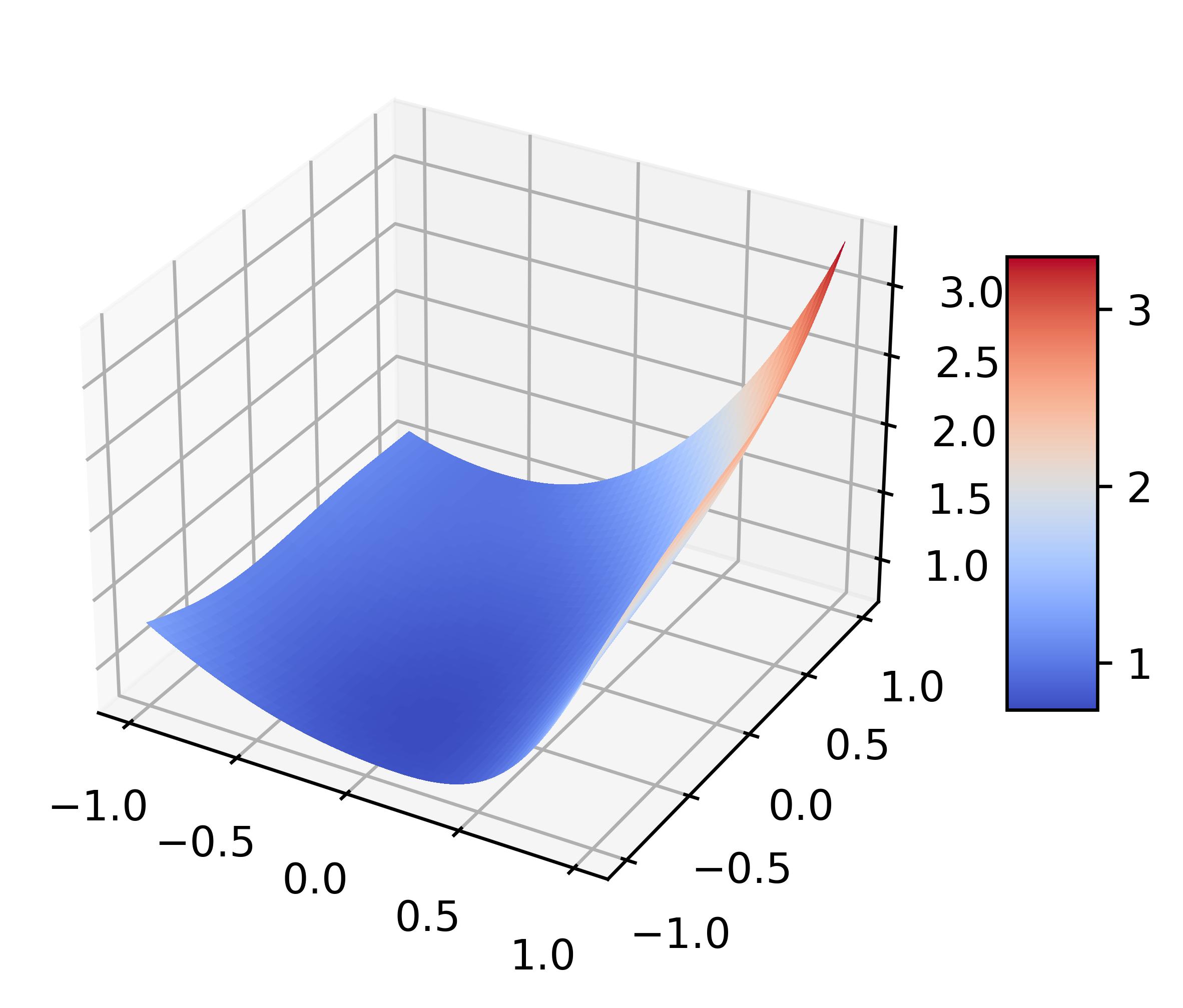}
    \label{fig:sub-first}}
    \hfill
    \subfloat[]{\includegraphics[width=0.3\linewidth]{ 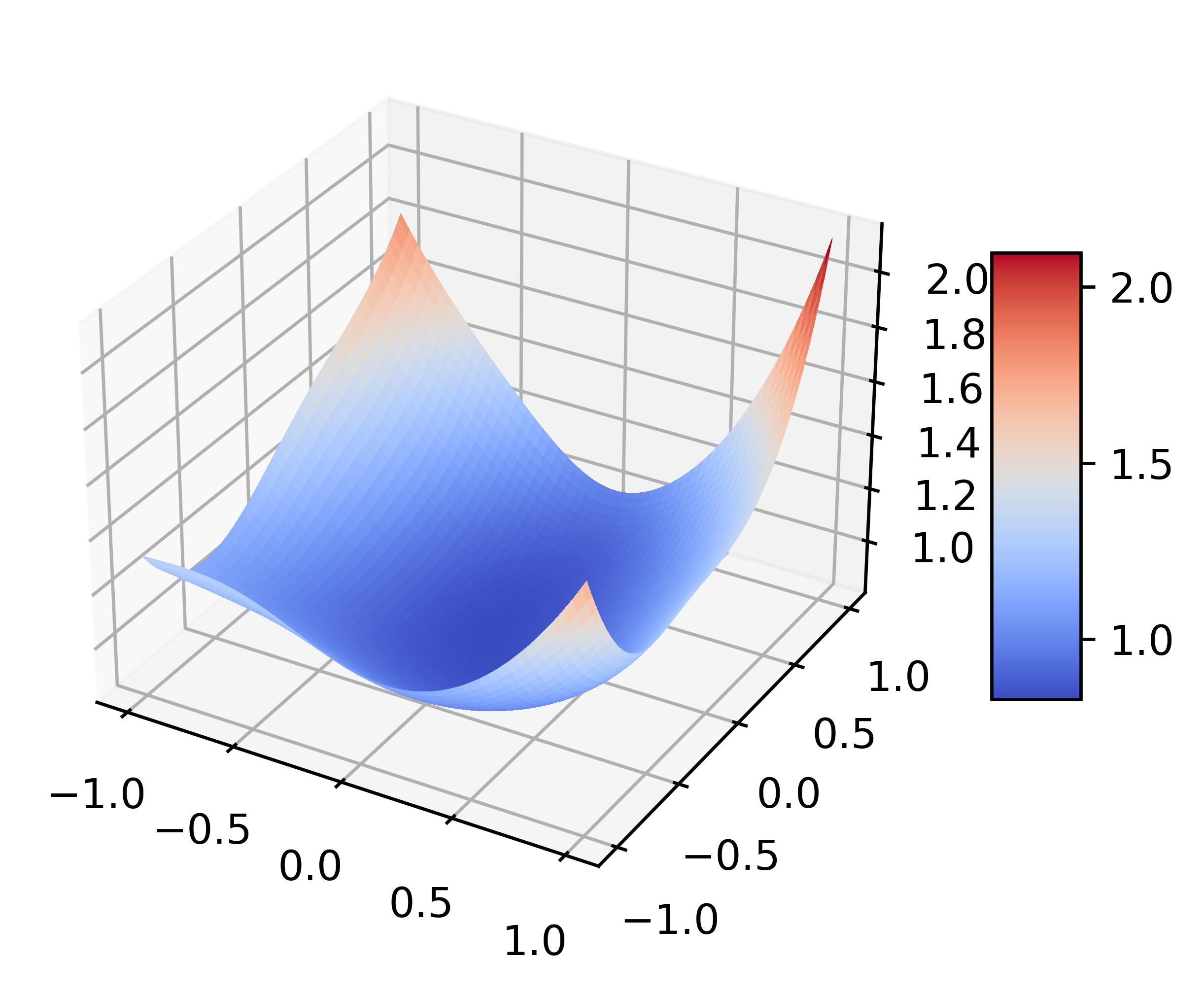}
    \label{fig:sub-first}}
\hfill
    \subfloat[]{\includegraphics[width=0.3\linewidth]{ 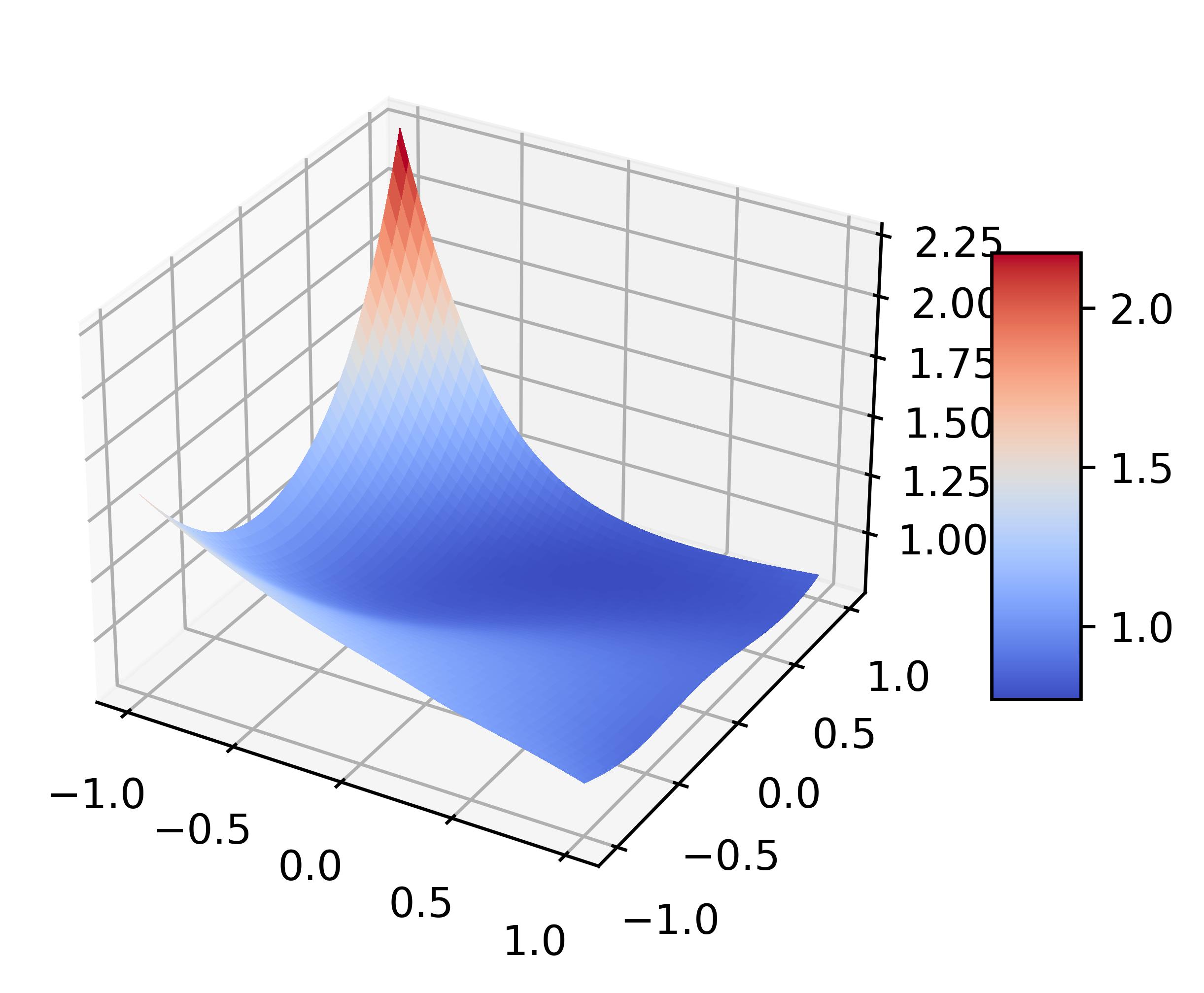}
    \label{fig:sub-first}}
    \subfloat[]{\includegraphics[width=0.3\linewidth]{ 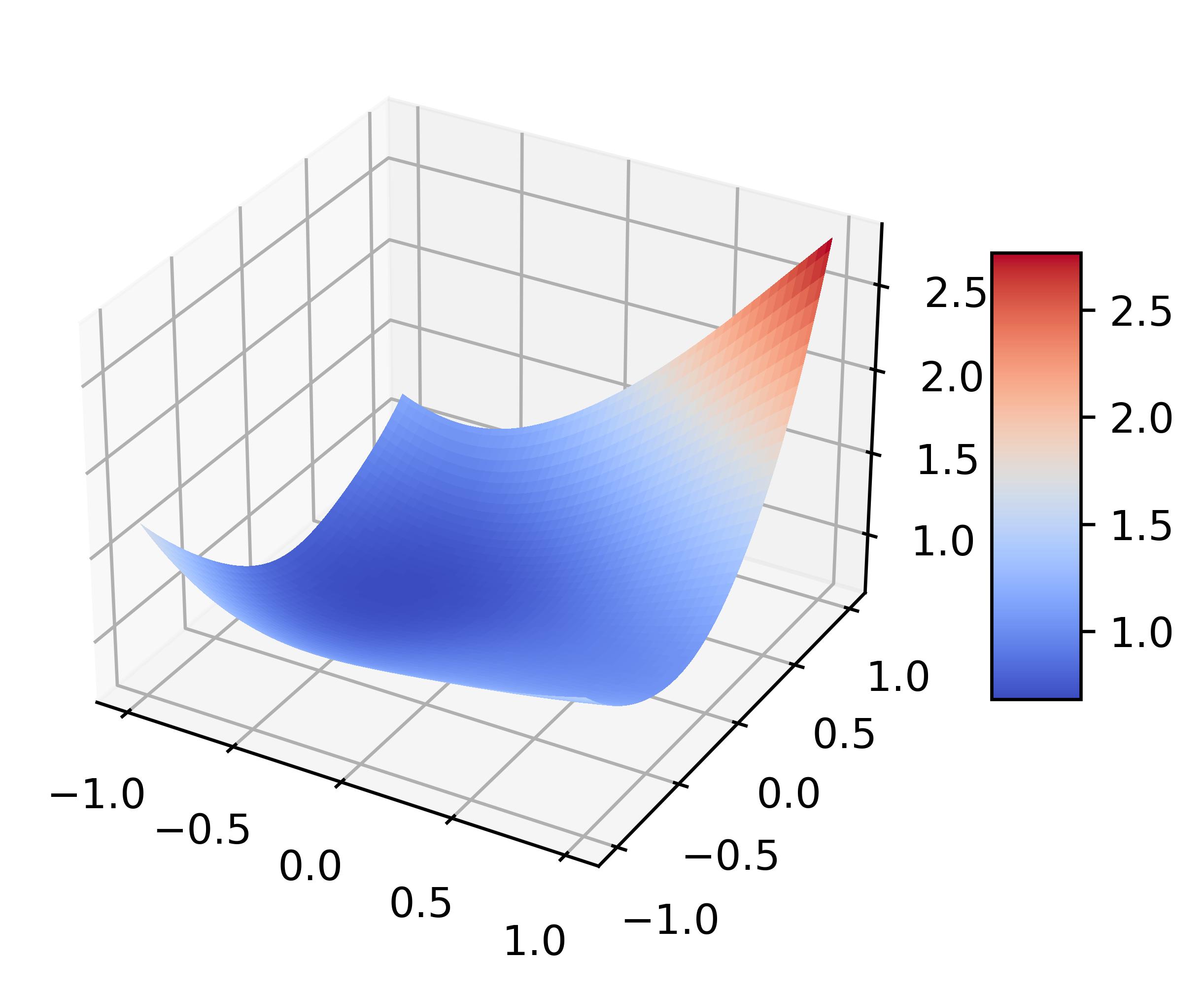}
    \label{fig:sub-first}}
\caption{(Best viewed in color) Output Landscape of a 4-Layer Dense Neural Network with ReLU and EELU Series. (a) ReLU, (b) EELU-1, (c) EELU-2, (d) EELU-3, (e) EELU-4 }
\label{fig:fig-6}
\end{figure*}
\subsection{Results of Activation Function Optimization Scheme}
\subsubsection{Superior Properties of the Developed Activation Functions}
Activation functions of the EELU series are non-linear, bounded below, and unbounded above, as shown in Figure \ref{fig:fig-5}(b) to Figure \ref{fig:fig-5}(e). Being bounded below helps in ensuring the strong regularization effects \citep{misra2019mish}. On the contrary, the unboundedness above is desired because it avoids saturation (significantly slow training due to near-zero gradients)~\citep{glorot2010understanding}. The negative bounds for the EELU series are approximately -0.3985, -0.3985, -0.1898, and -0.2755, respectively. Furthermore, the EELU series shows a non-monotonic bump in the positive $x$-axis region ($x>0$) for EELU-2 and a similar bump in the negative $x$-axis region ($x<0$) for the others. To be well-suited for gradient-based optimization, avoiding singularity (especially due to the lack of differentiability) is crucial. We investigated and found that each activation of the EELU series is continuously differentiable. {\color{myblue}In terms of smoothness and convergence, the EELU series shows better attributes than ReLU. A smooth transition in the output landscape effectively indicates a smoother loss landscape which eventually leads to a lower number of local minimums and faster convergence. To obtain the output landscape of the ReLU and EELU series, we employed a dense neural network of 4 layers following the work of \cite{visualloss}. These landscapes as shown in Figure \ref{fig:fig-6} are evident that the EELU series shows smoother transitions which can be a potential reason for the high performance of the EELU series in general.} A summary of the properties of the EELU series compared to the ReLU is demonstrated in Table~\ref{tab:table-properties}. All these attributes are evidence that the EELU series is suitable for image classification applications in deep neural networks.

\subsubsection{Performance on Different Neural Networks and Datasets}
The high performance of an activation function on a particular neural network with a specific dataset may not always secure the best performance in other architectures and datasets. To check the generalization over various neural architectures, we considered computationally heavy to light neural networks such as ResNet50~\citep{he2016deep}, AlexNet~\citep{krizhevsky2012imagenet}, VGG-16~\citep{simonyan2014very}, MobileNet~\citep{howard2017mobilenets}, and the base network $\phi$ using the CIFAR10 dataset. Then again, we tested the performance of the activation functions on different image classification datasets such as MNIST~\citep{lecun2010mnist}, Fashion MNIST~\citep{xiao2017fashion}, and Imagenette~\citep{imagenette} to validate the generalization over datasets. We used the base neural network $\phi$ mentioned in Table~\ref{tab:phi} for understanding the performance across different datasets. To check how the activation function performs in different domains such as agriculture, and medical science, we also tested the Beans dataset~\citep{beansdata} and Colorectal Histology dataset~\citep{kather2016multi} respectively. {\color{myblue}Furthermore, we evaluated the performance with large-scale benchmark datasets including TinyImageNet \citep{le2015tiny} and CottonWeedID15 \citep{chen2022performance} on ResNet50~\citep{he2016deep}, VGG-16~\citep{simonyan2014very}, and MobileNet~\citep{howard2017mobilenets}}. {\color{myblue}Finally, we compared the proposed activation functions using a Vision Transformer (ViT) model \citep{hassani2021escaping} with three recently proposed activation functions.} Therefore, our experiment included testing datasets of sizes ranging from 1,295 to 100,000 images. A summary of the datasets used in this study is outlined in \ref{sec:appendix_a}. All of these datasets are balanced (the number of images for each class is reasonably equal), except for the CottonWeedID15 \citep{chen2022performance}, and publicly available.
\begin{table*}[h]
\centering
\caption{Comparison of top-1 accuracy (mean with standard deviation) of activation functions in CIFAR10 dataset on different neural networks.}
\label{tab:table-6}
\resizebox{\linewidth}{!}{
\begin{tabular}{@{}ccccll@{}}
\hline
\begin{tabular}[c]{@{}c@{}}Activation Function\end{tabular} &
  Base CNN$(\phi)$  &
  ResNet50  &
  AlexNet &
VGG16 &
MobileNet \\ \hline
ReLU \citep{nair2010rectified}  & 81.37 ± 0.46          & 89.14 ± 0.28          & 77.25 ± 0.62          & 78.59 ± 0.29          & 69.85 ± 0.59 \\
Swish \citep{ramachandran2017searching} & 81.46 ± 0.60          & \textbf{90.15 ± 0.34} & 78.03 ± 0.33          & 77.46 ± 0.56          & 75.78 ± 0.81 \\
Mish \citep{misra2019mish}  & 82.23 ± 0.27          & 89.98 ± 0.25          & 78.01 ± 0.62          & 77.47 ± 0.14          & 76.70 ± 0.75 \\
GeLU \citep{hendrycks2016gaussian}  & 81.75 ± 0.26          & 89.63 ± 0.13          & 78.22 ± 0.51          & 77.04 ± 0.13          & 73.53 ± 1.08 \\
SERF \citep{nag2021serf}  & 82.16 ± 0.22          & 89.40 ± 0.17          & 78.12 ± 0.54          & 78.42 ± 0.11          & 75.54 ± 0.98 \\
ELU  \citep{clevert2015fast}  & 82.29 ± 0.57          & 89.92 ± 0.21          & 77.89 ± 0.41          & 78.76 ± 0.13          & \textbf{78.48 ± 2.23} \\ \hline
EELU-1 & 82.66 ± 0.29          & 89.91 ± 0.26          & 78.24 ± 0.47          & \textbf{79.02 ± 0.33} & 75.70 ± 0.46 \\
EELU-2 & 82.67 ± 0.30          & 89.96 ± 0.35          & \textbf{78.65 ± 0.42} & 78.72 ± 0.38          & 75.76 ± 0.58 \\
EELU-3 & 82.48 ± 0.61          & 89.65 ± 0.20          & 78.15 ± 0.28          & 78.60 ± 0.26          & 71.75 ± 0.50 \\
EELU-4 & \textbf{82.74 ± 0.14} & 89.75 ± 0.10          & 78.08 ± 0.47          & 78.59 ± 0.13          & 72.71 ± 1.23 \\ \hline
\end{tabular}}
\end{table*}

\begin{table*}[h]
\centering
\caption{Comparison of top-1 accuracy (mean with standard deviation) of activation functions in different datasets on the base network $\phi$.}
\label{tab:tab-d}
\resizebox{\linewidth}{!}{
\begin{tabular}{@{}ccccccc@{}}
\hline
\begin{tabular}[c]{@{}c@{}}Activation Function\end{tabular} &
  MNIST &
  \begin{tabular}[c]{@{}c@{}}Fashion MNIST \end{tabular} &
  CIFAR10  &
  Imagenette  &
  \begin{tabular}[c]{@{}c@{}}Colorectal Histology \end{tabular} &
  Beans \\ \hline
ReLU \citep{nair2010rectified}  & 99.23 ± 0.03          & 92.33 ± 0.24          & 81.37 ± 0.46          & 53.06 ± 0.73          & 70.31 ± 0.59          & 80.46 ± 0.23          \\
Swish \citep{ramachandran2017searching}  & 99.31 ± 0.04          & 92.00 ± 0.31          & 81.46 ± 0.60          & 55.71 ± 0.70          & 66.95 ± 0.52          & 82.19 ± 0.18          \\
Mish \citep{misra2019mish}   & 99.32 ± 0.03          & 92.46 ± 0.13          & 82.23 ± 0.27          & 57.38 ± 0.88          & 66.64 ± 0.55          & 80.16 ± 0.42          \\
GeLU \citep{hendrycks2016gaussian}   & 99.29 ± 0.08          & 92.10 ± 0.24          & 81.75 ± 0.26          & 57.42 ± 0.83          & 67.06 ± 0.73          & 81.40 ± 0.24          \\
SERF \citep{nag2021serf}   & 99.29 ± 0.05          & 92.56 ± 0.15          & 82.16 ± 0.22          & 57.02 ± 0.54          & 68.89 ± 0.68          & 81.25 ± 0.49          \\
ELU \citep{clevert2015fast}    & 99.30 ± 0.09          & 92.68 ± 0.09          & 82.29 ± 0.57          & 56.18 ± 0.51          & 66.65 ± 0.57          & 77.97 ± 0.03          \\ \hline
EELU-1 & 99.36 ± 0.05          & \textbf{92.79 ± 0.13} & 82.66 ± 0.29          & 57.66 ± 0.60          & 70.40 ± 0.62           & 82.43 ± 0.21          \\
EELU-2 & \textbf{99.38 ± 0.02} & 92.73 ± 0.11          & 82.67 ± 0.30          & 57.77 ± 0.78          & \textbf{73.45 ± 0.64} & 82.47 ± 0.36          \\
EELU-3 & 99.33 ± 0.07          & 92.70 ± 0.07          & 82.48 ± 0.61          & \textbf{59.24 ± 0.69} & 72.08 ± 0.63          & \textbf{82.65 ± 0.22} \\
EELU-4 & 99.35 ± 0.03          & 92.68 ± 0.25          & \textbf{82.74 ± 0.14} & 57.96 ± 0.83          & 71.42 ± 0.57          & 81.82 ± 0.16          \\\hline
\end{tabular}}
\end{table*}

{\color{myblue}
To ensure the fairness of the tests, we kept all other hyperparameters constant for each activation function. We evaluated the test accuracy over five repeated runs for each case (except for the compact convolutional transformer, which was tested for ten repetitions) and calculated the mean accuracy. Initially, we conducted tests on the CIFAR-10 dataset using the various neural networks mentioned earlier. The results, shown in Table~\ref{tab:table-6}, indicate that EELU-1, EELU-2, and EELU-4 outperform the other functions in VGG16, AlexNet, and the \( \phi \)-Network, respectively. However, for ResNet50 and MobileNet, Swish and ELU performed better, respectively. This discrepancy may be attributed to the architectural differences among the neural networks. AlexNet and VGG16 share a similar architecture to the \( \phi \) network used in AFOS, which may explain why our proposed framework produced activation functions that perform better in these networks. Conversely, ResNet50 and MobileNet incorporate unique features, such as skip connections and depth-wise separable convolutions, respectively. We hypothesize that using a base network with these features could enable AFOS to develop activation functions that perform optimally in these specialized networks.
}

\begin{table*}[h]
\centering
\caption{Comparison of top-1 accuracy (mean with standard deviation) of activation functions on AlexNet with different datasets}
\label{tab:alex}
\resizebox{\linewidth}{!}{
\begin{tabular}{@{}cccccc@{}}
\hline
Activation Function & MNIST      & Fashion MNIST        & Imagenette  & Beans            & Colorectal Histology\\ \hline
ReLU \citep{nair2010rectified}                & 99.01 ± 0.08 & 90.32 ± 0.16          & 66.01 ± 1.19 & 78.28 ± 3.46          & 81.13 ± 0.76         \\
Swish \citep{ramachandran2017searching}               & 99.05 ± 0.05 & 90.64 ± 0.05          & 66.16 ± 0.90 & 79.53 ± 1.94          & 83.77 ± 0.57         \\
Mish \citep{misra2019mish}                & 99.03 ± 0.08 & 90.65 ± 0.25          & 66.92 ± 2.77 & 80.46 ± 1.71          & 83.40 ± 0.32         \\
GeLU \citep{hendrycks2016gaussian}                 & 99.06 ± 0.04 & 90.65 ± 0.25          & 65.23 ± 2.05 & 79.68 ± 0.89          & 83.56 ± 0.77         \\
SERF \citep{nag2021serf}                & 99.05 ± 0.08 & 90.72 ± 0.03          & 66.37 ± 2.59 & 80.00 ± 2.95          & 83.74 ± 0.49         \\
ELU \citep{clevert2015fast}                & 99.04 ± 0.01 & 90.72 ± 0.13          & 64.15 ± 2.06 & 77.96 ± 3.76          & 82.40 ± 0.47         \\ \hline
EELU-1 & \textbf{99.07 ± 0.05} & 90.64 ± 0.19 & \textbf{68.01 ± 0.56} & 81.40 ± 2.56 & \textbf{84.16 ± 0.37} \\
EELU-2              & 99.05 ± 0.01 & \textbf{90.76 ± 0.31} & 66.75 ± 2.77 & 81.09 ± 0.91          & 83.89 ± 0.25         \\
EELU-3              & 99.03 ± 0.05 & 90.72 ± 0.43          & 67.23 ± 0.86 & 81.16 ± 1.41          & 83.19 ± 0.66         \\
EELU-4              & 99.05 ± 0.09 & 90.64 ± 0.49          & 67.80 ± 1.33 & \textbf{82.18 ± 1.56} & 84.10 ± 0.32         \\ \hline
\end{tabular}}
\end{table*}
\begin{table*}[h]
\centering
\caption{Comparison of top-1 accuracy (mean with standard deviation) of activation functions on VGG16 with different datasets}
\label{tab:vgg}
\resizebox{\linewidth}{!}{
\begin{tabular}{@{}cccccc@{}}
\hline
Activation Function & MNIST               & Fashion MNIST      & Imagenette         & Beans       & Colorectal Histology \\ \hline
ReLU \citep{nair2010rectified}               & 98.32 ± 0.16          & 91.36 ± 0.16          & 57.45 ± 4.23          & 75.78 ± 1.53 & 78.23 ± 1.55         \\
Swish \citep{ramachandran2017searching}               & 98.52 ± 0.23          & 91.11 ± 1.68          & 60.45 ± 2.13          & 79.69 ± 0.25 & 74.20 ± 5.79         \\
Mish \citep{misra2019mish}               & 98.71 ± 0.17          & 91.90 ± 0.25          & 58.39 ± 0.89          & 77.34 ± 2.06 & 78.89 ± 1.83         \\
GeLU \citep{hendrycks2016gaussian}                & 98.62 ± 0.06          & 90.48 ± 0.29          & 55.41 ± 1.99          & 79.69 ± 0.78 & 78.23 ± 3.13         \\
SERF \citep{nag2021serf}                & 98.67 ± 0.30          & 92.10 ± 0.19          & 62.59 ± 3.15          & 77.34 ± 1.85 & 77.06 ± 2.17         \\
ELU \citep{clevert2015fast}                & 98.38 ± 0.12          & 92.14 ± 0.05          & 65.98 ± 1.09          & 78.91 ± 2.21 & 74.16 ± 2.56         \\ \hline
EELU-1              & \textbf{98.97 ± 0.04} & 92.37 ± 0.13          & 66.19 ± 1.12          & 81.25 ± 0.17 & 79.26 ± 1.70         \\
EELU-2 & 98.81 ± 0.05 & 92.10 ± 0.15 & 66.64 ± 0.97 & \textbf{82.81 ± 0.92} & \textbf{80.03 ± 2.03} \\
EELU-3              & 98.72 ± 0.12          & \textbf{92.50 ± 0.23} & 61.49 ± 2.56          & 82.03 ± 1.63 & 79.53 ± 2.10         \\
EELU-4              & 98.61 ± 0.26          & 92.05 ± 0.10          & \textbf{66.76 ± 0.64} & 81.25 ± 1.28 & 79.26 ± 1.64         \\ \hline
\end{tabular}}
\end{table*}

{\color{myblue}Subsequently, we evaluated the performance of the developed activation functions on the MNIST, Fashion MNIST, Imagenette, Colorectal Histology, and Beans datasets. The results presented in Table~\ref{tab:tab-d} indicate that the EELU series consistently outperforms other activation functions across all these datasets. However, it is important to note that in these experiments, the base network \( \phi \) was kept constant. If the superior performance of the EELU series is maintained on AlexNet and VGG16 with different datasets, we can strongly recommend the EELU series for VGG-like architectures (such as the \( \phi \) Network, VGG-16, and AlexNet), regardless of the dataset used. Therefore, we conducted further tests on the AlexNet and VGG16 architectures using the aforementioned datasets. The results, as shown in Table~\ref{tab:alex} and Table~\ref{tab:vgg}, demonstrate that the EELU series dominates in every scenario. This indicates that the evolved activation functions excel in VGG-like neural networks and across a variety of image classification datasets. Consequently, we conclude that AFOS generates activation functions with robust generalization capabilities across diverse datasets and exhibits considerable generalization across different neural network architectures.}

\subsubsection{Statistical Comparison}
We conducted a Friedman Test~\citep{friedman1940comparison} to investigate the significance {\color{myblue}of differences in} the performance of the activation functions. {\color{myblue}The null hypothesis $H_0$ states that there is no difference between the performance of all activation functions, while the alternative hypothesis $H_A$ states that there is.} The Friedman statistics is calculated by the following equations  (\ref{eqn:eqn-4}) \& (\ref{eqn:eqn-3}):
\begin{equation}
      {R_i}= \frac{\sum_{q=1}^{N_M}r_{iq}}{N_M}
      \label{eqn:eqn-4}
\end{equation}
\begin{equation}
    {{\chi}^2}= \frac{12N_M}{N_A(N_A+1)}\left ( \sum_{i=1}^{N_A}{R_i}^2-\frac{N_A(N_A+1)^2}{4} \right )
    \label{eqn:eqn-3}
\end{equation}

where, $N_A$ denotes the number of activation functions compared, and $N_M$ represents the number of neural networks (or the number of datasets for comparing the performance among different datasets). In our study, $N_A$ and $N_M$ are 10 and 5 (16 for datasets), respectively. $R_i$ is the $i$-th activation function's average rank and $r_{iq}$ is the $i$-th activation function's rank on the $q$-th neural network (or dataset). For each neural network (or dataset), the activation functions were ranked from one to ten. A ranking of one represents the lowest accuracy, and conversely, ten represents the highest accuracy. Table~\ref{tab:table-9} and Table~\ref{tab:table-x} show the Friedman test results for accuracy at a significance level of 0.05. {\color{myblue}For large values of $N_A$ and $N_M$, Friedman statistics follows ${\chi}^2$ distribution \citep{iman1980approximations}, therefore we calculated the ${\chi}^2$ value for both neural network and datasets. In this study, the degree of freedom is $(N_M – 1)$ where $N_M$ is the number of individual cases considered for performance evaluation. Therefore, the degree of freedom is 4 and 15 for neural networks and datasets, respectively. The ${{\chi}^2}_{Critical}$ values for neural networks and datasets were approximately 9.49 and 25.00, respectively. Since the ${{\chi}^2}_{Critical}$ value is lower than ${\chi}^2$ statistics in both cases, we reject the null hypothesis. Therefore, we conclude that performance is not the same for all ten activation functions. Additionally, higher rank and significant difference in the performance indicate the superior performance of the proposed activation function series.} The average rank ($R_i$) of EELU-2 is the highest in terms of different neural networks and different datasets among all ten activation functions, {\color{myblue}which ultimately indicates the superiority of EELU-2 over the state-of-the-art activation functions. The properties of EELU-2 such as non-linearity, differentiability, non-monotonicity, and lower-boundedness are similar to the other activation functions of the EELU series. The only dissimilarity, a rather unique property of EELU-2 is in the direction of the upper bound, which is negative. Even with this unique characteristic, EELU-2 outperformed the state-of-the-art activation functions in almost all cases.

\begin{table}[h]
\centering
\caption{Result of Friedman test based on different neural network}
\label{tab:table-9}
\resizebox{0.7\linewidth}{!}{
\begin{tabular}{@{}cccccc@{}}
\hline
\begin{tabular}[c]{@{}c@{}}Activation Function\end{tabular} & \begin{tabular}[c]{@{}c@{}}Network Based \\Rank $R_i$\end{tabular}  & ${\chi}^2$                    & p-value                          & {\color{myblue}\begin{tabular}[c]{@{}c@{}}Hypothesis\\Test\\Result\end{tabular}}                  \\ \hline
ReLU \citep{nair2010rectified}                                                         & 2.0   & \multirow{10}{*}{17.86} &
\multirow{10}{*}
{\textless{}0.05} &
\multirow{10}{*}{{\color{myblue}Reject $H_0$}} \\  \cline{1-2}
Swish \citep{ramachandran2017searching}   & 5.2                  &  &  \\ \cline{1-2}
Mish \citep{misra2019mish}  & 5.8                 &  &  \\ \cline{1-2}
GeLU \citep{hendrycks2016gaussian}   & 3.8                 &  &  \\  \cline{1-2}
SERF \citep{nag2021serf}   & 4.2                  &  &  \\\cline{1-2}
ELU \citep{clevert2015fast}   & 6.8                & &  \\ \cline{1-2}
EELU-1 & 7.8                 &  &  \\ \cline{1-2}
EELU-2 & \textbf{8.4}  &    &  \\  \cline{1-2}
EELU-3 & 5.4                 &  &  \\ \cline{1-2}
EELU-4 & 5.6           &    &  \\ \hline
\end{tabular}}
\end{table}

Another observation from the Friedman test is about the generalization of the performance of the EELU series. The ${\chi}^2$ value for the dataset-based ranking is more significant than that of the neural network-based ranking. This is because the $p$-value associated with the ${\chi}^2$ value of 81.84 is less than 0.001, while the $p$-value associated with the ${\chi}^2$ value of 17.86 is significant but not as strong, as it is greater than 0.001 but less than 0.05. This lower $p$-value indicates the consistent performance of the EELU series across different datasets, in other words, `generalization'. Conversely, the generalization across different neural networks is less strong compared to different datasets.} 

\begin{table}[h]
\centering
\caption{Result of Friedman test based on different dataset}
\label{tab:table-x}
\resizebox{0.7\linewidth}{!}{
\begin{tabular}{@{}cccccc@{}}
\hline
\begin{tabular}[c]{@{}c@{}}Activation Function\end{tabular} &  \begin{tabular}[c]{@{}c@{}}Dataset Based\\Rank $R_i$\end{tabular} & ${\chi}^2$                    & p-value                          & {\color{myblue}\begin{tabular}[c]{@{}c@{}}Hypothesis\\Test\\Result\end{tabular}}                  \\ \hline
ReLU \citep{nair2010rectified}                                                        & 2.4 &\multirow{10}{*}{81.84} &
\multirow{10}{*}{\textless{}0.05} &
\multirow{10}{*}{{\color{myblue}Reject $H_0$}} \\ \cline{1-2}
Swish \citep{ramachandran2017searching}           & 4.0          &  &  &  \\\cline{1-2}
Mish \citep{misra2019mish}           & 4.5          &  &  &  \\ \cline{1-2}
GeLU \citep{hendrycks2016gaussian}         & 4.3          &  &  &  \\\cline{1-2}
SERF \citep{nag2021serf}         & 5.1          &  &  &  \\\cline{1-2}
ELU \citep{clevert2015fast}         & 3.8          & & &  \\\cline{1-2}
EELU-1         & 8.5          &  &  &  \\ \cline{1-2}
EELU-2 & \textbf{8.7} &  &  &  \\ \cline{1-2}
EELU-3         & 7.6          &  &  &  \\\cline{1-2}
EELU-4         & 7.5          &  &  &  \\ \hline
\end{tabular}}
\end{table}

{\color{myblue}
\subsubsection{Tests on Large-scale Benchmark Datasets}To evaluate the effectiveness of the EELU series on large-scale and challenging datasets, we conducted experiments with the TinyImageNet benchmark dataset (200 classes, balanced) \citep{le2015tiny} and the CottonWeedID15 dataset (15 classes, imbalanced) \citep{chen2022performance} as shown in Table \ref{tab:tiny}. For the CottonWeedID15 dataset \citep{chen2022performance}, we performed 5-fold cross-validation since it does not have any specified test set. Moreover, since this dataset has class imbalance i.e., the number of images in each class is not the same, we reported the mean and standard deviation of the F1-scores in Table \ref{tab:tiny}. For the TinyImageNet dataset \citep{le2015tiny}, we trained the respective models with the training set and then evaluated the performance on the validation set. Due to the longer training time and limited resources, we evaluated the ReLU activation function as the baseline along with the EELU series. 
\begin{table*}[h]
\centering

\caption{Comparison of performance of ReLU and EELU series across three different neural architectures with CottonWeedID15 and TinyImageNet. The mean and standard deviation of the F1-score from five-fold cross-validation is reported for CottonWeedID15. For TinyImageNet, the mean and standard deviation of five repeated runs have been reported.}
\label{tab:tiny}

\resizebox{\linewidth}{!}{
\begin{tabular}{@{}ccccccc@{}}
\hline
Dataset                     & \multicolumn{3}{c}{CottonWeedID15}                           & \multicolumn{3}{c}{TinyImageNet} \\ \hline
\multicolumn{1}{c|}{Neural Network} & ResNet50 & VGG16 & \multicolumn{1}{c|}{MobileNet} & ResNet50 & VGG16 & MobileNet \\ \hline
\multicolumn{1}{c|}{ReLU}   & 89.32 ± 1.63 & 88.53 ± 1.17 & \multicolumn{1}{c|}{77.30 ± 1.52} &     51.34 ± 0.94    &    48.17 ± 0.63      &    45.78 ± 0.53      \\
\multicolumn{1}{c|}{EELU-1} & 90.24 ± 2.08 & 89.27 ± 1.62 & \multicolumn{1}{c|}{\textbf{83.13 ± 1.80}}&      \textbf{52.61 ± 0.87}     &     49.83 ± 0.85      &    \textbf{46.92 ± 0.47}      \\
\multicolumn{1}{c|}{EELU-2} & \textbf{91.13 ± 2.19} & \textbf{90.73 ± 1.29} & \multicolumn{1}{c|}{82.90 ± 1.86} &    52.56 ± 1.04       &  \textbf{49.92 ± 1.04}       &    46.38 ± 0.82      \\
\multicolumn{1}{c|}{EELU-3} & 91.08 ± 1.18 & 88.87 ± 1.74 & \multicolumn{1}{c|}{79.85 ± 2.45} &     51.87 ± 1.13      &    48.76 ± 0.78       &     46.28 ± 0.36     \\
\multicolumn{1}{c|}{EELU-4} & 90.89 ± 2.63 & 89.90 ± 1.31 & \multicolumn{1}{c|}{80.94 ± 2.23} &     52.23 ± 0.68      &    49.21 ± 0.93       & 45.96 ± 1.06         \\ \hline
\end{tabular}%
}
\end{table*}

For each of these datasets, we have used the weight of models pre-trained on the ImageNet dataset \citep{ILSVRC15}. Table \ref{tab:tiny} demonstrates the superior performance of the EELU series, especially EELU-1, and EELU-2 compared to ReLU. ImageNet \citep{ILSVRC15} is the most widely used benchmark dataset for image classification; however, due to its large size and our limited computational resources, we were not able to test our proposed activation functions on this dataset. However, we believe that the performance of the EELU series on the datasets we used is indicative of its potential effectiveness on larger benchmark datasets. Tests on the CottonWeedID15 and TinyImageNet were not included in the Friedman test because we only evaluated ReLU as the baseline for these experiments.}

{\color{myblue}
\subsubsection{Tests on Vision Transformer}
Vision transformers \citep{dosovitskiy2020image} are widely used unique computer vision models with high performance. Therefore, we also tested the performance of our developed activation functions on the compact convolutional transformer (CCT) \citep{hassani2021escaping} using the CIFAR10 dataset. Table \ref{tab:cct} demonstrates that our proposed EELU activation functions still outperform the standard and new baseline activation functions. For this comparison, we have introduced three new activation functions that were proposed recently, including NIPUNA \citep{madhu2023nipuna}, SupEx \citep{kiliccarslan2023detection}, and Smish \citep{wang2022smish}. Test on the CCT was not included in the Friedman test because we evaluated three additional activation functions as the baseline for this experiment.}
\begin{table}[ht]
    \centering
    \caption{\color{myblue}{Top-1 accuracy of different activation functions on compact convolutional transformer \citep{hassani2021escaping} using the CIFAR10 dataset. Mean and standard deviations of ten replications have been reported.}}
    \label{tab:cct}
    \color{myblue}{
    \resizebox{0.6\linewidth}{!}{
    \begin{tabular}{cc}
        \hline
        \textbf{Activation Function} & \textbf{Accuracy} \\ \hline
        ReLU \citep{nair2010rectified} & 80.54 ± 0.37 \\ 
        Swish \citep{ramachandran2017searching} & 80.75 ± 0.43  \\ 
        Mish \citep{misra2019mish} & 80.81 ± 0.42 \\ 
        GeLU \citep{hendrycks2016gaussian}& 80.40 ± 0.52 \\ 
        SERF \citep{nag2021serf}& 80.76 ± 0.63  \\ 
        ELU \citep{clevert2015fast}& 80.48 ± 0.66 \\ 
        NIPUNA \citep{madhu2023nipuna} & 80.55 ± 0.54  \\ 
        SupEx \citep{kiliccarslan2023detection} & 79.96 ± 0.66 \\ 
        Smish \citep{wang2022smish} & 81.13  ± 0.41 \\ \hline
        EELU-1 & \textbf{81.35 ± 0.50} \\ 
        EELU-2 & 81.22 ± 0.56 \\ 
        EELU-3 & 81.19 ± 0.48 \\ 
        EELU-4 & 81.08 ± 0.60 \\ \hline
    \end{tabular}
    }
    }
    
\end{table}
\subsubsection{{\color{myblue}Comparison of Computation Time}}
{\color{myblue}

In our implementation, we utilized the \verb|tensorflow.math.erf(x)| function provided by TensorFlow\footnote{\url{https://www.tensorflow.org/api_docs/python/tf/math/erf}}, which is an optimized and efficient implementation (includes GPU support) of the Gaussian error function. Tensorflow uses polynomial and rational approximation to calculate the function value. This approach ensured that the computational overhead was minimized while maintaining the accuracy of the activation function.

To assess the computational overhead introduced by the \texttt{erf} operation in the EELU activation function, we conducted an experiment to measure the average time taken to compute the \texttt{erf} function. Using a random array of 1000 values, we applied the \texttt{erf} function across 10,000 runs and recorded the total computation time. The experiment was conducted on GPU NVIDIA GeForce RTX 2080 SUPER on Intel Core i9-10900, 32.0 GB RAM, and 64-bit Windows 10 Operating System.

Our findings indicate that the average time to compute the \texttt{erf} operation is approximately $1.17 \times 10^{-5}$ seconds per run. This minimal computational overhead demonstrates that the inclusion of the \texttt{erf} operation in the EELU activation function does not significantly impact the overall training time. We conducted additional experiments to measure the computational time required by the EELU activation function, specifically focusing on the impact of the \texttt{erf} operation.

Our experiments compared the computational time of EELU with other standard activation functions. We measured the time taken per forward and backward pass during training on the same hardware configuration for the Fashion MNIST dataset on ResNet50 architecture. The results indicated that while the \texttt{erf} operation in EELU introduces a slight increase in computational time, the overhead is minimal and does not significantly affect the overall training time.

The results in Table~\ref{tab:table-10} show that EELU-3 is the second most efficient in terms of computational cost. While it is slightly more demanding than ReLU, it outperforms ReLU on all tested neural networks (Table~\ref{tab:table-9}) and datasets (Table~\ref{tab:table-x}). EELU-1 also demonstrates competitive computational efficiency, being faster than Mish, GeLU, and SERF. It ranks the second in performance across various neural networks and datasets. Although EELU-2 has a higher computational cost compared to EELU-1 and EELU-3, it remains faster than GeLU and SERF. Moreover, EELU-2 shows outstanding performance and ranks the highest across different datasets and neural networks.}

\begin{table*}[h]
\centering
\caption{Comparison of average time taken per step with ResNet50 and Fashion MNIST dataset}
\label{tab:table-10}
\resizebox{\linewidth}{!}{
\begin{tabular}{@{}ccccccccccc@{}}
\hline
\begin{tabular}[c]{@{}c@{}}Activation Function\end{tabular} &
  ReLU &
  Swish  &
  Mish &
  GeLU&
  SERF&
  ELU &
  EELU-1 &
  EELU-2 &
  EELU-3 &
  EELU-4 \\ \hline
\begin{tabular}[c]{@{}c@{}}Time/step (ms)\end{tabular} &
  \textbf{27.78} &
  29.02 &
  32.97 &
  34.76 &
  37.22 &
  \textbf{28.84} &
  30.86 &
  34.06 &
  \textbf{28.28} &
  35.49 \\ \hline
\end{tabular}}
\end{table*}
\section{Conclusions and Future Work}
\label{sec:conclusions}
{\color{black}
In this study, we proposed an activation function optimization scheme for image classification problems. With an enhanced search space and an efficient fitness function, we optimized and developed a series of activation functions named as EELU series. These activation functions possess the desired properties {\color{myblue}such as} non-linearity, lower-boundedness, differentiability, smoothness, and non-monotonicity. Comparative experimental results validate the superior performance of the EELU series both across different neural networks and across different datasets. Finally, the Friedman test conforms the statistical significance of the performance of the EELU series from the other standard activation functions. Even with the negligible trade-off of computation time per step, the improved accuracy and strong generalization over various datasets and neural networks justify that the activation functions of the EELU Series are robust choices for solving image classification problems through deep learning. However, we conclude that the generalization of the developed activation functions over a variety of datasets is stronger than the generalization over different neural networks. {\color{myblue}In the future, one potential research focus could involve exploring the optimization of activation functions using a specifically designed neural network. This network would aim to ensure robust generalization across a diverse range of neural networks. Another avenue for investigation would be to analyze how other hyperparameters affect the performance of the EELU series and the convergence dynamics of these activation functions. Moreover, it would be insightful to examine the performance of the proposed activation functions in other tasks, such as language translation, speech recognition, time series analysis, and more.} 

\section*{CRediT Author Statement}
\textbf{Abdur Rahman:} Conceptualization, Methodology, Software, Formal analysis, Writing - Original Draft, Visualization. \textbf{Lu He:} Conceptualization, Methodology, Writing - Review \& Editing. \textbf{Haifeng Wang:} Conceptualization, Methodology, Writing - Review \& Editing.

\section*{Declaration of Competing Interest}
The authors declare that they have no known competing financial interests or personal relationships that could have appeared to influence the work reported in this paper.

\section*{Data Availability}
The datasets used in this study are publicly available.

\section*{Acknowledgements}
This research did not receive any specific grant from funding agencies in the public, commercial, or not-for-profit sectors.
\bibliographystyle{elsarticle-harv} 
\bibliography{cas-refs}
\clearpage





\appendix

\section{Parameters used in AFOS}
\label{sec:appendix_c}

\setcounter{table}{0}
\begin{table}[h]
\centering
\caption{Parameters used in AFOS}
\label{tab:table-2}
\resizebox{0.4\linewidth}{!}{
\begin{tabular}{@{}ll@{}}
\hline
\textbf{Parameters}          & \textbf{Values/Options}   \\ \hline
$J$                & 30                        \\
$N$         & 40 \\
$S$                      & Ranking selection         \\
$X$                & Single point crossover \\
$x$ & 80\%                      \\
$\tau$ & Random                      \\
$M$                      & Single point mutation           \\
$m$           & 20\%                       \\
$n_{\Psi}$   & 6\\
$max\;  n_x$ & 6\\
$max\;  n_m$ & 1\\
$min\; n_{rand}$ & 17\\
$D_{Tr}$ & 45,000\\
$D_V$ & 5,000\\
$D_{Te}$ & 10,000\\\hline
\end{tabular}%
}
\end{table}
\clearpage
\section{Datasets}
\label{sec:appendix_a}
\setcounter{table}{0}
\begin{table}[h]
\centering
\caption{{\color{myblue}Datasets Used to Study the Comparison of Activation Functions}}
\label{tab:table-4}
\resizebox{\linewidth}{!}{
{\color{myblue}
\begin{tabular}{@{}cccccc@{}}
\hline
Dataset                & Number of Classes & Training Images & Test Images & Image shape & Class Balancing\\ \hline
MNIST \citep{lecun2010mnist}  & 10      & 60,000         & 10,000     & 28×28×1     & Balanced \\
Fashion MNIST \citep{xiao2017fashion}        & 10      & 60,000         & 10,000     & 28×28×1     & Balanced \\
CIFAR10 \citep{Krizhevsky09learningmultiple}             & 10      & 50,000         & 10,000     & 32×32×3     & Balanced \\
Imagenette \citep{imagenette}            & 10      & 9,469          & 3,925      & 160×160×3   & Balanced \\
Colorectal   Histology \citep{kather2016multi} & 8       & 3,750          & 1,250      & 150×150×3   & Balanced \\
Beans \citep{beansdata}              & 3       & 1,034          & 261        & 500×500×3   & Balanced \\
CottonWeedID15 \citep{chen2022performance}  & 15       & 5,187          & -        & 128×128×3   & Imbalanced \\
TinyImageNet \citep{le2015tiny}   & 200       & 100,000          & 10,000       & 64×64×3   & Balanced \\\hline
\end{tabular}}}
\end{table}

\end{document}